\pdfoutput=1
\documentclass[11pt]{article}
\usepackage[table]{xcolor}
\usepackage[colorinlistoftodos]{todonotes}
\usepackage[final]{acl}

\usepackage{fancyhdr}
\fancypagestyle{firstpage}{
  \fancyhf{}
    \fancyhead[L]{\normalsize Accepted in EMNLP 2025}
     \fancyfoot[C]{\thepage}
}
\usepackage{times}
\usepackage{latexsym}

\usepackage[T1]{fontenc} 
\usepackage[utf8]{inputenc}

\usepackage{microtype}

\usepackage{inconsolata}

\usepackage{graphicx}
\usepackage{booktabs}
\usepackage{amsmath}
\usepackage{multirow}
\usepackage{array}
\usepackage{tabularx}
\usepackage{arydshln} % For dashed lines
\usepackage{caption}
\usepackage{hyperref}

\title{PCRI: Measuring Context Robustness in Multimodal Models for Enterprise Applications}

\author{
\textbf{Hitesh Laxmichand Patel\textsuperscript{}},
\textbf{Amit Agarwal\textsuperscript{}},
\textbf{Srikant Panda\textsuperscript{}},
\textbf{Hansa Meghwani\textsuperscript{}},
\textbf{Karan Dua\textsuperscript{}}, \\
\textbf{Paul Li\textsuperscript{}}, 
\textbf{Tao Sheng\textsuperscript{}},
\textbf{Sujith Ravi\textsuperscript{}},
\textbf{Dan Roth\textsuperscript{}}
\\
\\
 \textsuperscript{}Oracle AI
\\
 \small{
   \textbf{Correspondence:} \href{mailto:email@domain}{hitesh.laxmichand.patel@oracle.com}
 }
}

\begin{document}
\maketitle
\thispagestyle{firstpage}
\pagestyle{firstpage}
% \begin{abstract}
% Multimodal Large Language Models (MLLMs) heavily rely on attention mechanisms to solve vision-language tasks, but their performance often declines in complex or noisy visual contexts. To systematically investigate this issue, we introduce the Patch Context Robustness Index (PCRI), a novel score quantifying how effectively MLLMs handle varying visual context granularities. PCRI uniquely measures sensitivity as the performance difference between localized image patches and full-image contexts, explicitly capturing a model's robustness to visual distractions. Benchmarking 19 state-of-the-art models on 15 diverse vision-language datasets, we find consistent improvements when models use localized patches, indicating substantial sensitivity to visual noise. Models employing hierarchical attention mechanisms exhibit significantly greater robustness. PCRI provides researchers and practitioners a standardized evaluation tool to quantify context sensitivity, guiding adaptive visual encoding and attention mechanism design toward robust, real-world multimodal applications.
% \end{abstract}

\begin{abstract}
The reliability of Multimodal Large Language Models (MLLMs) in real-world settings is often undermined by sensitivity to irrelevant or distracting visual context, an aspect not captured by existing evaluation metrics. We introduce the \textbf{Patch Context Robustness Index (PCRI)}, the first systematic and interpretable score for quantifying MLLM robustness to variations in visual context granularity, measuring performance changes between localized image patches and full-image input.

Applying PCRI to 19 state-of-the-art MLLMs across 15 vision-language benchmarks, we find that most leading models remain brittle to background noise, with only a few, such as InternVL2-26B and Qwen2VL-72B, demonstrating consistent robustness across tasks. PCRI analysis also highlights how different model architectures handle and integrate visual context, offering actionable diagnostic insight for both researchers and practitioners.

% PCRI analysis also reveals which tasks and architectures benefit most from global versus local context, offering actionable diagnostic insight for both researchers and practitioners.

PCRI enables rigorous comparison of context robustness, % across models and tasks, 
supporting  principled model selection and guiding the development of future architectures and training strategies for robust, real-world deployment.
\end{abstract}

\section{Introduction}

MLLMs have rapidly transformed real-world applications such as visual question answering \cite{pattnayak2024survey,pattnayak2025hybrid}, e-commerce product search \cite{meghwani-etal-2025-hard}, interactive assistants, document understanding,  \cite{patel2024llm,patel2025sweeval,agarwal-etal-2025-fs}, accessibility for visually impaired users \cite{panda2025whosaskinginvestigatingbias,panda2025daiqauditingdemographicattribute},  and synthetic data-pipelines \cite{agarwal2024techniques,agarwal2024synthetic}. In these deployments, models must
reliably extract relevant cues from complex visual scenes, for example, correctly identifying a product despite background clutter or assisting visually impaired users in noisy environments, to support safety, fairness, \& user experience.
However, despite impressive progress in academic benchmarks, current MLLMs often fail to generalize when exposed to complex, noisy, or dynamic visual environments.%CAMERA encountered in practice.

Current evaluation protocols typically measure model performance on static, full-image contexts, implicitly assuming uniform relevance of all visual regions or relying on model's capability to filter the relevant information to solve a given task. This assumption rarely holds in practice: real-world images often contain clutter, occlusions, or irrelevant backgrounds that can mislead even advanced models. In contrast, we explicitly evaluate model behavior on both full images and localized image patches to examine the sensitivity to visual context granularity. Recent studies have documented failures due to missed local details ~\cite{zhang2024exploringperceptuallimitationmultimodal,zhang2023small}, fragmentation from cropping~\cite{zhu2024hyvilm,ma2024infllava}, and performance drops under visual perturbations~\cite{qiu2024benchmarkingrobustnessmultimodalimagetext}. Such failures have direct implications for real-world reliability, user trust, and downstream decision making.

Practitioners and system designers need tools to quantify and compare the context robustness of MLLMs, defined as the ability of the models to maintain performance when the visual scene changes in granularity or distractor content, enabling informed model selection, and mitigation of hidden failure modes. However, to our knowledge, no standardized metric or score currently exists to quantify this form of robustness.

In this work, we introduce Patch Context Robustness Index (PCRI), a novel practical score to measure the sensitivity of MLLMs to variations in visual context granularity. PCRI directly quantifies performance differences when models process localized image patches versus full-image contexts. %,CAMERA, capturing the degree to which models are robust or brittle to visual distractions and noise. 
Our contributions are as follows:

\begin{itemize}
    \item We propose PCRI, the first quantitative score designed specifically to measure the context robustness of MLLMs under varying visual granularities.
    \item We present a structured, patch-based evaluation framework to systematically understand sensitivity to visual context in MLLMs.
     \item We present a large-scale study of 19 state-of-the-art MLLMs on 15 vision-language datasets, revealing significant and previously unmeasured context sensitivity.
\end{itemize}

% By establishing PCRI as a standardized, interpretable score, we enable both researchers and practitioners to rigorously evaluate, compare, and select MLLMs fit for real-world deployment, and further improve development of these models.

Our evaluation shows that even leading MLLMs remain surprisingly brittle to context variation, with only a few architectures demonstrating robust, human-like reasoning.

\section{Related Work}

\textbf{Robustness Benchmarks for Multimodal Models.}
The increasing adoption of MLLMs in real-world applications has driven extensive research on their robustness to input variations. Recent works have constructed challenging benchmarks to probe model reliability under diverse perturbations \cite{agarwal-etal-2025-mvtamperbench}. \citet{qiu2024benchmarkingrobustnessmultimodalimagetext} (MMRobustness) systematically evaluate MLLMs on distribution shifts via 17 image and 16 text perturbation techniques, introducing metrics like MultiModal Impact (MMI) and Missing Object Rate (MOR). Similarly, R-Bench~\cite{li2025r} targets real-world corruptions by modeling the complete imaging pipeline, including in-the-wild and machine-induced distortions across 33 dimensions, and proposes comprehensive robustness evaluations for 20 MLLMs. Both studies highlight the vulnerability of MLLMs to common and complex perturbations, yet focus primarily on distribution shift and absolute/relative performance drops under corruptions.

\noindent \textbf{Task-Specific and Contextual Robustness.}
Beyond generic robustness, certain tasks probe more nuanced forms of context sensitivity. For example, VCR~\cite{zhang2025vcr} challenges vision-language models to restore occluded embedded text, requiring pixel-level reasoning about local and global context. While such tasks advance the frontier of context-aware modeling, their evaluations are task-specific and do not yield general-purpose robustness metrics or score. %CAMERA for arbitrary vision-language scenarios.

\noindent \textbf{Attention Mechanisms and Visual Context in MLLMs.}
Numerous studies investigate the failure modes of attention mechanisms in MLLMs, revealing sensitivity to object size, distractors, and spatial arrangement~\cite{zhang2024exploringperceptuallimitationmultimodal,zhang2023small}. Architectural innovations, such as multi-resolution encoding~\cite{ma2024infllava, zhu2024hyvilm, thapa2024dragonfly}, token pruning~\cite{chen2024vltpvisionlanguageguidedtoken}, and text-relevant patch selection~\cite{ye2024efficient} have been developed to mitigate context distractions, but typically optimize for efficiency or accuracy without providing standardized, interpretable measures of context robustness.

\textbf{Summary and Our Contribution.}
In summary, while recent benchmarks and metrics have significantly advanced the evaluation of MLLMs under distributional shift, corruption, and task-specific complexity, there remains a critical gap: \textit{no prior work provides a unified, score-driven framework for quantifying MLLM robustness to visual context granularity across diverse tasks and architectures}. PCRI fills this gap, offering a standardized, interpretable, and broadly applicable score for evaluating and comparing the context robustness of MLLMs, and enabling more principled model selection and deployment in practice.

\begin{figure*}[th!]
    \centering
    \includegraphics[width=\textwidth]{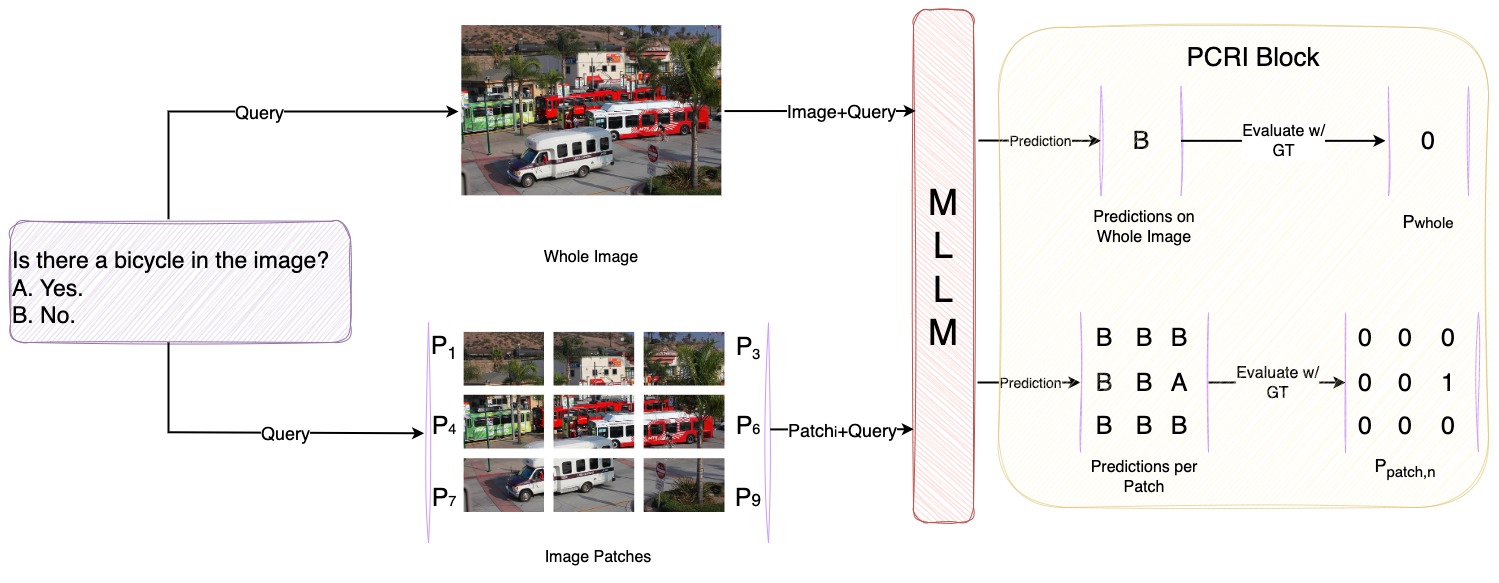}
    \caption{PCRI-based evaluation framework. An MLLM processes a query using either the full image (top) or individual patches (bottom). Predictions are compared against ground truth to compute PCRI, assessing robustness to context variations in multimodal reasoning.}
    \label{fig:tampered_frame_image}
    \vspace{-1.5em}
\end{figure*}

\section{Methodology}
\label{sec:methodology}

Our goal is to systematically evaluate whether MLLMs can reason robustly over both localized and global visual contexts, an essential capability for real-world deployment, where distractors and irrelevant background are common. Existing evaluation protocols rarely measure this aspect, and ad hoc approaches (such as object-centric cropping) introduce biases and are impractical at scale. We therefore introduce a simple, reproducible patch-based framework to quantify context robustness, suitable for diverse models and tasks.

\paragraph{Patch-Based Evaluation Framework.}
Given an image, we partition it into $n \times n$ non-overlapping, equally sized patches, with $n$ controlling the granularity. For each patch, the model is evaluated independently, isolating its ability to extract information from other patches%localized context
 (see Figure~\ref{fig:tampered_frame_image}). A regular grid ensures unbiased, interpretable, and systematic analysis, avoiding the pitfalls of object-centric or saliency-based methods, and enabling direct comparison across models and datasets. We evaluate at three granularities:
\begin{itemize}
    \item \textbf{Full-Image Context ($n=1$):} The model receives the entire image.
    \item \textbf{Moderate Granularity ($n=2$):} The model receives each $2 \times 2$ patch independently.
    \item \textbf{Fine Granularity ($n=3$):} The model receives each $3 \times 3$ patch independently.
\end{itemize}
Larger $n$ were explored in ablations (Appendix \ref{sec:appendix_granular} \& ~\ref{sec:appendix_granularity}), with diminishing returns and substantially increased computational cost ($n^2$ evaluations per image).

\paragraph{Patch Context Robustness Index (PCRI).}
PCRI quantifies a model's sensitivity to changes in visual context granularity. Formally, for a given $n$:
\begin{equation}
    \text{PCRI}_n = 1 - \frac{P_{\text{patch}, n}}{P_{\text{whole}}}
    \label{eq:pcri}
\end{equation}
where:
\begin{itemize}
    \item $P_{\text{patch}, n}$ is the maximum performance achieved (per sample) over all $n \times n$ patches.
    % \item $P_{\text{whole}}$ is the performance when the full image is provided.
    \item $P_{\text{whole}}$ is the performance with the full image.% is provided.
\end{itemize}
% We use \textit{maximum} patch score to capture the best-case model behavior, ensuring the metric reflects the model's potential to leverage any local cue, rather than averaging out successes and failures (see Appendix~\ref{sec:PCRI_properties} for further discussion).

% \paragraph{Aggregation policy (max over patches).}
% We aggregate patch scores with a \emph{max} operator to probe best-case local solvability: if any localized region suffices, a robust model should exploit it. Because patches present minimal contextual noise (e.g., background clutter or layout distractors), selecting the best patch against the full image more cleanly reveals whether performance derives from local evidence or requires global context. Averaging would dilute informative regions with many irrelevant ones and understate sensitivity to global context (see App.~\ref{sec:PCRI_properties}).

\paragraph{Aggregation policy (max over patches).}
We aggregate patch scores with a \emph{max} operator because PCRI diagnoses whether global context distracts the model. Patches provide minimal-context views; contrasting the best local patch with the full image reveals if access to global context helps or hurts. If the best patch rivals or exceeds full-image performance, the model is likely relying on spurious global cues (global context as a distractor); if it lags, the task or model benefits from global integration. Averaging over patches would dilute informative regions with many irrelevant ones and obscure this distraction signal (see Appendix~\ref{sec:PCRI_properties}).

% \paragraph{Aggregation policy (max over patches).}
% We aggregate patch scores with a \emph{max} operator to probe best-case local solvability: if any localized region suffices, a robust model should exploit it. Because patches offer minimal contextual noise (e.g., background clutter or layout distractors), selecting the best patch against the full image more cleanly reveals whether performance derives from local evidence or requires global context. Averaging would dilute informative regions with many irrelevant ones and understate sensitivity to global context (see App.~\ref{sec:PCRI_properties}).

% \paragraph{Aggregation policy (max over patches).}
% We aggregate patch scores with a \emph{max} operator to probe best-case local solvability: if any localized region suffices, a robust model should exploit it. Averaging would dilute informative regions with many irrelevant ones and understate sensitivity to global context (see App.~\ref{sec:PCRI_properties}).
% We use the \emph{maximum} patch score to probe best-case local solvability: if any localized region suffices, a robust model should capitalize on it without global context. Averaging would dilute informative patches with non-informative ones and understate sensitivity to global context (see App.~\ref{sec:PCRI_properties}).

\paragraph{Interpretation.}
% CAMERA PCRI is a comparative score, not a normalized score or metric, capturing how sensitive model performance is to visual context granularity, while being agnostic to any metric and dataset. 
PCRI is a comparative score, agnostic to indiviudal metrics and dataset, that captures a model’s sensitivity to visual context granularity. Table~\ref{tab:pcri-interpretation} summarizes the key scenarios.% For tasks that require global context within the image to solve, it is expected the performance drop at patch level, while tasks that cannot be addressed with whole image lead to undefined scores.

\begin{table}[h]
    \centering
    \scalebox{0.92}{
    \begin{tabular}{p{2.6cm}p{5cm}}
    \toprule
    \textbf{PCRI Value} & \textbf{Interpretation} \\
    \midrule
    $\approx 0$ & Model is robust; performs equally well on full image and patches. \\
    $< 0$ & Global context distracts; model harmed by irrelevant background. \\
    $> 0,\ \leq 1$ & Model needs global context to solve the task; patch input omits necessary information. \\
    $\ll 0$ or undefined & $P_{\text{whole}} \to 0$; model cannot solve task even on the full image—interpret with caution. \\
    \bottomrule
    \end{tabular}}
    \caption{Summary of PCRI score interpretation. Each PCRI range indicates a distinct model behavior with respect to robustness against visual context.}
    \label{tab:pcri-interpretation}
    \vspace{-1em}
\end{table}

\paragraph{Validity Domain \& Chance.}
We interpret PCRI only when the full-image score is meaningfully above the dataset-specific chance floor. Let $C(d)$ denote the chance level for dataset $d$ (e.g., $1/|\mathcal{Y}|$ for balanced $|\mathcal{Y}|$-way classification, or the documented random baseline for retrieval/captioning metrics). A model-dataset pair $(m,d)$ is considered \emph{valid} if:
\[
\begin{aligned}
P_{\text{whole}}(d,m) &\ge C(d)+\Delta_{\min},\\
\Delta_{\min} &= \max\{\delta,\,2\,\mathrm{SE}\}.
\end{aligned}
\]

% Here $\mathrm{SE}$ is the standard error of $P_{\text{whole}}$ estimated via non-parametric bootstrap over evaluation examples, and $\delta$ is a small absolute margin on the native scale of the base metric. Unless otherwise stated, we use $\delta=0.01$ on $[0,1]$-scaled metrics (i.e., $1.0$ percentage point). Pairs ($(m,d)$) failing this gate are labeled \emph{near-chance/unstable}, and practioioners should not compute compute or interpret PCRI for such pairs. In our experiments, all model–dataset pairs satisfied this gate; consequently, we report and interpret PCRI for all results.

% Here, $\mathrm{SE}$ is the standard error of $P_{\text{whole}}$, estimated via nonparametric bootstrap over evaluation examples, and $\delta$ is a small absolute margin on the native scale of the base metric. Unless otherwise stated, we set $\delta=0.01$ for $[0,1]$-scaled metrics (i.e., 1.0 percentage point). Model–dataset pairs $(m,d)$ that fail this gate are labeled \emph{near-chance/unstable}; for such pairs, practitioners should not compute or interpret PCRI. In our experiments, all model–dataset pairs satisfy this gate; therefore, we report and interpret PCRI for all results.

\begin{figure*}[!ht]
    \centering
    \includegraphics[width=\linewidth]{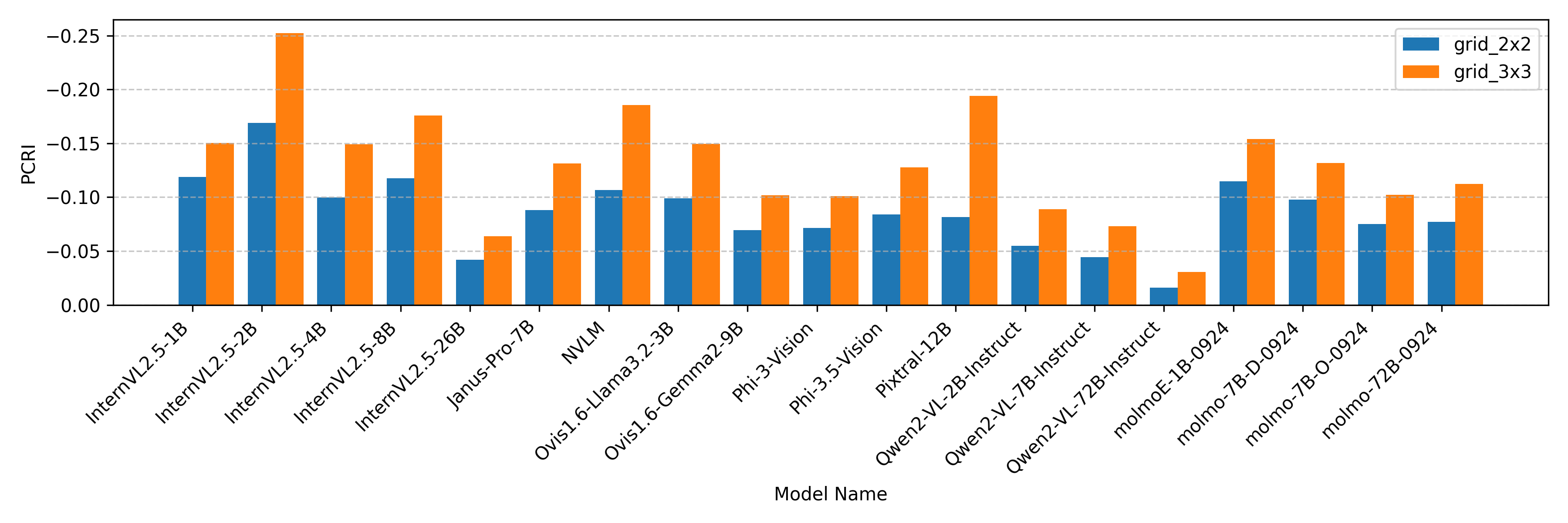}
        
    \caption{Avg. PCRI across 15 benchmarks for 19 MLLMs at \(2 \times 2\) and \(3 \times 3\) granularities. Lower PCRI values highlight model succeeds on local patches but fails on the whole image highlighting the sensitivity of the models.}
    \vspace{-1em}
    \label{fig:avg_avg}
\end{figure*}

Here, $\mathrm{SE}$ is the standard error of $P_{\text{whole}}$, estimated via nonparametric bootstrap over evaluation examples, and $\delta$ is a small absolute margin on the native scale of the base metric. Unless otherwise stated, we set $\delta=0.01$ for $[0,1]$-scaled metrics (i.e., 1.0 percentage point). Model–dataset pairs $(m,d)$ that fail this gate are labeled \emph{near-chance/unstable}; in such cases, practitioners should not compute or interpret PCRI and may report only the underlying task metrics. In our experiments, all model–dataset pairs satisfy this gate; therefore, we report and interpret PCRI for all results. Further details can be found in Appendix \ref{sec:chance-validity}.

% \paragraph{Chance-normalized PCRI (cPCRI).}
% Let $FIP(d,m)$ denote the full-image performance of model $m$ on dataset $d$, and let $Patch(d,m)$ denote the performance with $n$-patch inputs under our max-over-patches policy. Let $\chance(d)$ denote the dataset-specific chance level for the task/metric (e.g., $1/|\mathcal{Y}|$ for balanced $|\mathcal{Y}|$-way classification, or a documented random-baseline for retrieval/captioning metrics). We define a companion, chance-normalized metric:
% \[
% \text{cPCRI}_n(d,m)\;=\;1 \;-\; \frac{\PPatch(d,m) - \chance(d)}{\FIP(d,m) - \chance(d)}\quad\text{(when }\FIP(d,m)>\chance(d)\text{).}
% \]
% Intuitively, $\text{cPCRI}_n$ rescales patch performance relative to the gap between full-image performance and chance, improving cross-task comparability. As with PCRI, lower values indicate greater reliance on global context: if patches alone cannot recover full-image performance, $\text{cPCRI}_n$ increases.

\section{Experiments \& Results}

% We apply our PCRI-based evaluation across a comprehensive range of models \& datasets, ensuring our evaluation captures variations across task types, model architectures, and training paradigms to systematically assess the robustness of MLLMs under varying visual context granularities. 

% [CAMERA] We systematically evaluate PCRI across a diverse set of models and datasets, capturing variations in task type, model architecture, and training paradigm. This enables a comprehensive assessment of MLLM robustness to changes in visual context granularity.

We evaluate PCRI on diverse models and datasets to comprehensively assess MLLM robustness to visual context granularity.

\noindent \textbf{Benchmarks.} We evaluate across multiple %vision-language 
benchmarks by categorizing them into type of tasks:% as follows:

 - Image Captioning: MS-COCO Captions

 - Multiple-Choice QA (MCQ): AI2D, BLINK, MMMU, MMStar, RealWorldQA, ScienceQA
 
 - Yes/No Classification: AMBER, HallusionBench, MME, POPE

 - Visual Question Answering (VQA): ChartQA, GQA, TextVQA, VizWiz

\paragraph{Models.} We benchmark 19 state-of-the-art MLLMs, across various model family and sizes. 

% \paragraph{Granularity and compute.}
% Unless noted otherwise, we use small grids $n\in\{2,3\}$ (4 or 9 patches) for the best insight-per-compute. Evaluation cost grows with the number of patches relative to a single full-image pass; batching amortizes I/O but not model FLOPs. Finer grids ($n\!\ge\!4$) increasingly fragment semantically coherent evidence (objects, text lines, part–whole relations), which inflates false negatives for local solvability and blurs the “global-context distraction” signal that PCRI is designed to measure. In practice, we recommend $n{=}2$ for fast gating/regression checks and $n{=}3$ on a stratified subset when deeper diagnostics are required. Sensitivity checks indicate conclusions are stable across $n\in\{2,3\}$ (see App.~\ref{app:granularity}).

\paragraph{Granularity and Compute.}
We default to small grids $n\in\{2,3\}$ (4–9 patches) for the best insight-per-compute; evaluation cost scales with $n^2$ relative to a single full-image pass. Larger $n$ tends to fragment coherent evidence and weakens PCRI’s global-context distraction probe (see Appendix~\ref{sec:PCRI_properties}). For consistency \& reproducibility, all evaluations use VLMEvalKit~\cite{duan2024vlmevalkit}. Details of the datasets \& models is in Appendix~\ref{sec:data_models}.%are provided in Appendix~\ref{sec:data_models}.

% \begin{itemize}
%     \item {Multiple-Choice QA (MCQ)}: AI2D, ScienceQA, RealWorldQA
%     \item {Yes/No Classification}: AMBER, POPE, HallusionBench
%     \item {Visual Question Answering (VQA)}: GQA, ChartQA, TextVQA, VizWiz   
%     \item {Image Captioning}: MS-COCO Captions
% \end{itemize}

% For consistency and reproducibility, evaluations are conducted using VLMEvalKit \cite{duan2024vlmevalkit}. We provide a detailed discussion of the types of dataset and the models benchmarked in the Appendix \ref{sec:data_models}.

\subsection{Results \& Discussion}

\begin{figure*}[!ht]
    \centering
    \includegraphics[width=\linewidth]{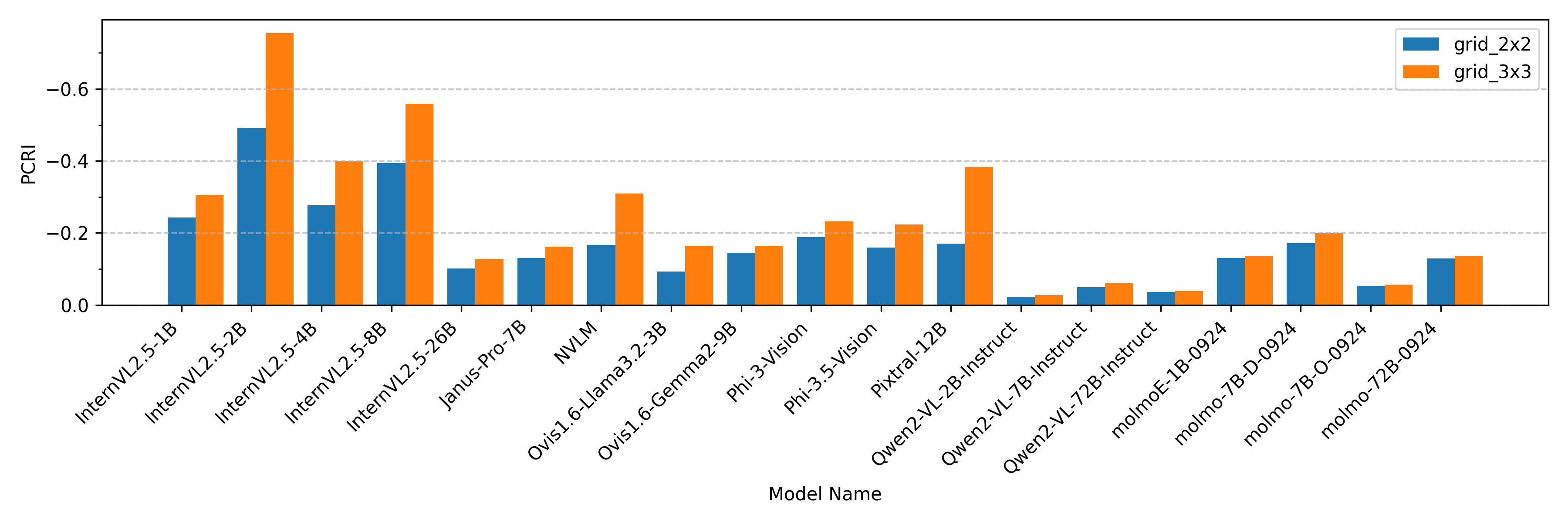}
    
    \caption{Avg. PCRI on MS-COCO Captioning task for different MLLMs at \(2 \times 2\) and \(3 \times 3\) granularities. Lower PCRI values highlight stronger performance on localized patches versus full-image contexts, notably in smaller InternVL2.5  models (<26B), NVLM and Pixtral models.}
    \vspace{-1.2em}
    \label{fig:caption_avg}
\end{figure*}

\begin{figure*}[!ht]
    \centering
    \includegraphics[width=\linewidth]{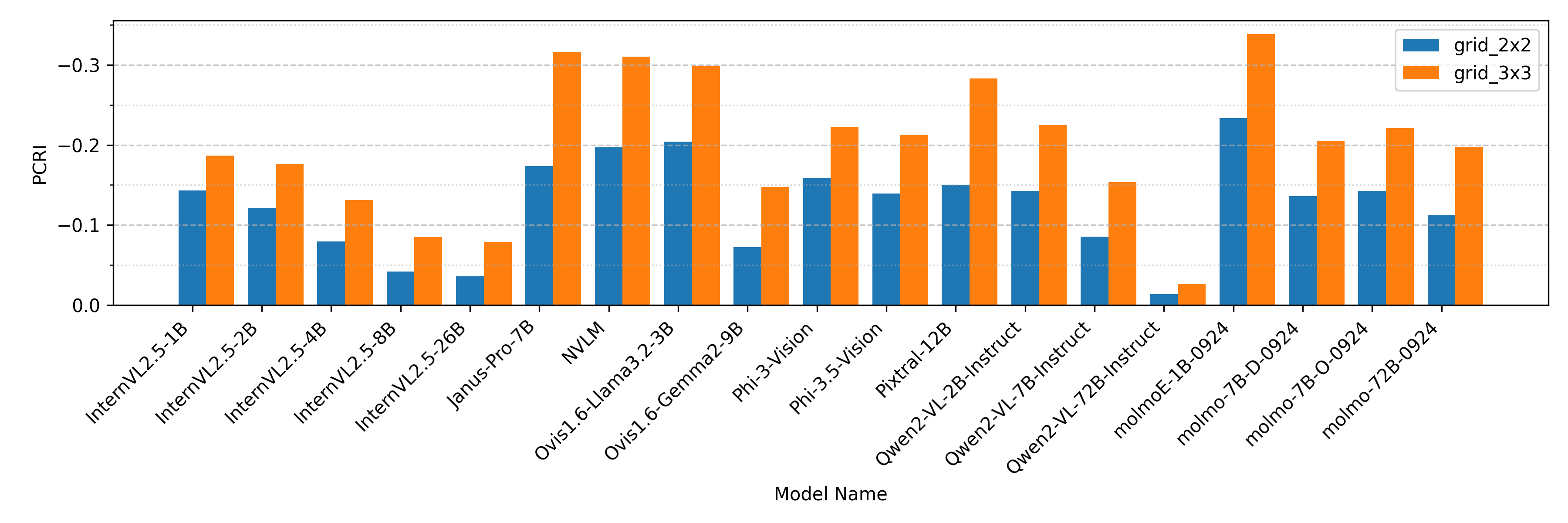}
    \caption{Avg. PCRI scores for MCQ tasks. Models exhibit moderate sensitivity to visual context granularity, with larger models (e.g., Qwen2-VL-72B) demonstrating enhanced robustness to global contextual noise, with consistent improved performance with finer patch granularities (3×3) highlights ongoing benefits of localized visual cues.}
    \vspace{-1em}
    \label{fig:mcq_avg}
\end{figure*}

We organize our analysis around core research questions central to evaluating PCRI’s validity, utility \& context robustness of MLLMs. Each subsection directly addresses one of these questions.

\subsubsection{Do MLLMs Favor Localized Patches over Full Images?}

The majority of the 19 benchmarked MLLMs exhibit negative PCRI values (Figure~\ref{fig:avg_avg}), indicating better performance on localized patches than on full images. For most models, global visual context introduces noise or distraction that outweighs its benefits for task performance. This pattern is especially pronounced in smaller InternVL variants, NVLM, and Pixtral, which show the lowest PCRI values, suggesting greater sensitivity to irrelevant context as model size or alignment decreases. Possible contributors to this trend include:
\vspace{-0.3em}
\begin{itemize}
    \item \textbf{Visual Distraction:} Difficulty in filtering %irrelevant 
    background reduces accuracy on full images.
    \item \textbf{Cross-Modal Attention Misalignment:} Suboptimal alignment with text prompts leads to diluted focus in global contexts.
    \item \textbf{Attention Overload:} Increased tokens in full images can 
    % \item \textbf{Attention Overload:} Increased token count in full images %can 
    overwhelm attention mechanisms.% CAMERA should cite in final version ~\cite{chen2024vltpvisionlanguageguidedtoken, zhang2024redundancy}.
\end{itemize}

% CAMERA DETAILED
% \begin{itemize}
%     \item \textbf{Visual Noise and Distraction:} Many models fail to suppress irrelevant information in the full image (global context), leading to performance drop relative to focused patches.
%     \item \textbf{Cross-Modal Attention Misalignment:} Visual features may not be sufficiently aligned with task-relevant textual cues, diluting the model’s focus in global contexts.
%     \item \textbf{Attention Overload:} Larger context windows (full images) dramatically increase token count, straining attention mechanisms and impairing reasoning. % CAMERA should cite in final version ~\cite{chen2024vltpvisionlanguageguidedtoken, zhang2024redundancy}.
% \end{itemize}

In contrast, models such as InternVL2-26B \& Qwen2VL-72B display near-zero PCRI, reflecting comparatively higher robustness to context variation \& improved integration of global context. Overall, while a few models are closing the gap, most current MLLMs remain sensitive to context.% granularity.
\vspace{-0.5em}
\subsubsection{Does PCRI Align with Human-Like Contextual Reasoning?}

To validate PCRI’s interpretability, we conducted a human study in which three annotators solved vision-language tasks across two datasets, using both individual patches and the full image (see Appendix~\ref{sec:human_study_appendix} for details). Humans always performed as well or better with the full image; performance never exceeded the patch-only condition. This robust pattern of context use contrasts with MLLMs showing negative PCRI, where patch-only input outperforms the full image: a deviation from human-like reasoning that signals shortcut exploitation or context brittleness. PCRI thus not only quantifies robustness, but also highlights departures from desirable, human-style contextual reasoning.

% \subsubsection{Does PCRI align with Human-like Contextual Reasoning?}

% To validate the interpretability of PCRI, we conducted a human study in which annotators (3 authors) solved vision-language tasks across two datasets, using both individual image patches and the full image. Human performance always improved or remained the same when provided the full image, never exceeding patch-only performance, reflecting natural human reasoning. In contrast, we find that models with negative PCRI (higher patch performance than full image) are exploiting shortcuts or suffering from context brittleness, a behavior not aligned with human cognition. Thus, PCRI not only quantifies model robustness but also reveals deviations from desirable, human-like reasoning. Full protocol and results are in Appendix~\ref{sec:human_study_appendix}.

% % \subsubsection{How do different vision-language tasks influence context sensitivity?}

\begin{figure*}[!ht]
    \centering
    \includegraphics[width=\linewidth]{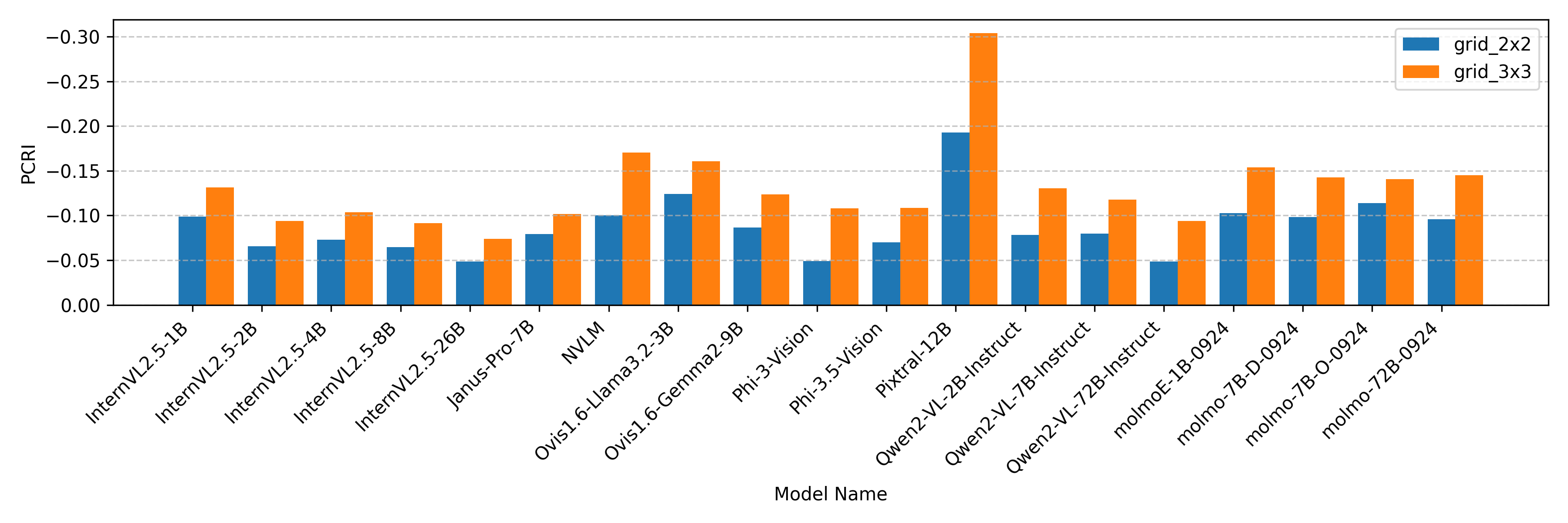}
    \caption{Avg. PCRI scores for Yes/No tasks across evaluated MLLMs. Most MLLMs show lower context sensitivity in binary decision-making scenarios compared to captioning tasks, but consistently improved performance with finer patch granularities (3×3) highlights ongoing benefits of localized visual cues.}
    % }

    \vspace{-1em}
    \label{fig:yn_avg}
\end{figure*}

\subsubsection{How does Task-Type modulate Context Sensitivity?}% in vision-language models?}
% CAMERA We observe clear differences in PCRI across task types, reflecting varying reliance on contextual information:

% PCRI varies across task types, reflecting their differing reliance on contextual information:

\textbf{Captioning Tasks:}
Captioning (Figure \ref{fig:caption_avg}) consistently show the strongest negative PCRI values, notably with InternVL2.5 (up to -0.49 at $2\times2$ and -0.75 at $3\times3$ for 2B). % [CAMERA READY}This aligns with prior evidence~\cite{anderson2018bottomup}, suggesting 
This suggests that image captioning inherently focuses on localized entities rather than global scene understanding, making these models particularly susceptible to background objects and noise. 

\textbf{Multiple-Choice QA (MCQ) Tasks:}
MCQ tasks (Figure \ref{fig:mcq_avg}) also exhibit negative PCRI scores, but with moderate sensitivity compared to captioning. Models such as Qwen2-VL and InternVL perform relatively better, supporting prior claims that MCQ tasks often leverage textual biases or selective visual attention mechanisms~\cite{agrawal2016analyzing}. However, models like Janus-Pro-7B and NVLM suffer significantly more, likely due to less sophisticated visual encoding strategies. Further details are in Appendix \ref{sec:appendix_mcq}.

\textbf{Yes/No Classification Tasks:}
Yes/No tasks (Figure \ref{fig:yn_avg}) exhibit the mildest PCRI scores, indicating that binary visual reasoning typically involves simpler or fewer visual cues, reducing the dependency on full-image context. Nevertheless, notable exceptions such as Pixtral-12B (-0.30 at $3\times3$) highlight significant variability and sensitivity, suggesting that model-specific factors such as visual encoder design % (ROPE-2D encoding) 
affecting task robustness more than task complexity alone. Further details are in Appendix \ref{sec:appendix_yn} \& Appendix \ref{sec:appendix_vqa} for VQA tasks.

\subsubsection{Are all Models equally Context Sensitive?}
% CAMERA Model-specific PCRI patterns provide deeper insights into design trade-offs and robustness behaviors across different tasks. For further details see Appendix \ref{sec:appendix_model_level_analysis}.

Model-specific PCRI patterns reveal design trade-offs and robustness behaviors; see Appendix~\ref{sec:appendix_model_level_analysis} for details.

\paragraph{InternVL Models:}
InternVL variants exhibit varied context sensitivity across tasks. Strongly negative PCRI scores in captioning indicate that these models excel at fine-grained object recognition, yet struggle with holistic scene reasoning. Conversely, InternVL's relatively better context robustness on MCQ and Yes/No tasks likely stems from its dynamic resolution mechanisms and hierarchical attention layers, which facilitate effective selective encoding% should cite in final version ~\cite{chen2024internvl}
. Notably, the larger variant (26B) demonstrates superior resilience, validating the efficacy of hierarchical attention at larger scales.

\paragraph{Molmo Models:}
Molmo demonstrates relatively consistent PCRI values across tasks, highlighting strong robustness due to effective cross-modal alignment strategies. However, despite stable context robustness, its absolute performance is moderate in tasks demanding detailed visual reasoning (MCQ and captioning). This suggests that Molmo achieves robustness through generalized visual-textual alignment but at the cost of specialization for context-sensitive tasks.% should cite in final version ~\cite{deitke2024molmo}.

\paragraph{Qwen2-VL Models:}
Qwen2-VL models show pronounced negative PCRI trends, particularly at lower scales% uncomment in final version (2B and 7B)
, emphasizing their reliance on localized visual recognition strategies established through contrastive pre-training methods. The largest Qwen2-VL (72B) model notably achieves better robustness, likely benefiting from advanced multimodal rotary positional embeddings (M-RoPE) and resolution-adaptive encoding strategies, enhancing its global context integration capability. % should cite in final version ~\cite{wang2024qwen2}.

\subsubsection{How does image granularity ($n=2,3$) impact Model Robustness?}
Increasing the granularity from $2\times2$ to $3\times3$ consistently amplifies negative PCRI scores across tasks (Figures \ref{fig:caption_avg},\ref{fig:mcq_avg},\ref{fig:yn_avg}), indicating that finer granularities further improve localized context performance relative to full-image contexts. Notably, larger models like 72B \& 26B % uncomment in final version Qwen2-VL-72B \& InternVL2.5-26B 
exhibit relatively lower PCRI magnitude shifts, suggesting that larger-scale hierarchical attention mechanisms provide enhanced robustness against visual context shifts. See Appendix~\ref{sec:appendix_granularity} for additional details and Appendix~\ref{sec:example_PCRI} for qualitative examples.

\subsection{Implications for Industry and Research}

PCRI offers a rigorous, interpretable tool for evaluating MLLMs under real-world deployment constraints. Its key applications include:

\begin{itemize}
    \item \textbf{Model Selection and Qualification:} PCRI provides a standardized criterion to identify \& select models that maintain performance under variable or noisy visual conditions, supporting safer deployment in high-stakes domains (e.g., %accessibility,
    content moderation).
    
    \item \textbf{Model Design and Diagnosis:} By revealing context brittleness and patch sensitivity, PCRI pin points architectural weaknesses and guides targeted improvements, such as enhancing hierarchical attention, integrating retrieval-augmented modules, or optimizing cross-modal alignment.%CAMERA, beyond what aggregate accuracy can reveal.
    
    % [CAMERA] \item \textbf{Dataset Curation and Benchmarking:} PCRI can highlight datasets or benchmarks where context granularity is most challenging, guiding dataset augmentation or annotation efforts to address hidden brittleness.
    
    \item \textbf{Continuous Monitoring and Auditing:} In production, PCRI enables ongoing tracking of context robustness as data evolves, facilitating early detection of emerging vulnerabilities, crucial for regulatory compliance, user safety, and long-term reliability.
    
    %{CAMERA} \item \textbf{Model Documentation and Transparency:} PCRI scores can be included in model cards or documentation to communicate context robustness to downstream users and stakeholders, aiding responsible AI deployment.
\end{itemize}

For practitioners, if PCRI is strongly negative, further investigation or model retraining is recommended; if near zero, the model can be trusted to generalize across visual contexts. See Appendix~\ref{sec:practitioner_appendix} for practical interpretation \& real-world examples.

\section{Conclusion}

We introduce the Patch Context Robustness Index (PCRI), a scalable \& interpretable score for systematically quantifying the sensitivity of MLLMs to changes in visual context granularity. Our evaluation, spanning 19 recent MLLMs \& 15 diverse tasks, provides the first comprehensive study %
% , model-agnostic benchmark 
of context brittleness in the field.

Our analysis reveals that most MLLMs remain vulnerable to irrelevant or distracting context, with negative PCRI scores indicating performance degradation in the presence of full-scene information. In contrast, models such as InternVL2-26B and Qwen2VL-72B demonstrate superior context robustness across benchmarks, providing actionable choices for practitioners facing real-world noise and clutter. We also find substantial variation across task types, highlighting where global or local context is most essential.

% PCRI enables direct comparison of model robustness, beyond aggregate accuracy—
PCRI enables direct comparison of model robustness, supporting both diagnostic evaluation \& production deployment decisions. We demonstrate that prioritizing models with near-zero PCRI in our beta rollout led to measurably better user experience and reliability, even where task-level accuracy was matched.

By establishing a unified and extensible evaluation framework, our work lays the foundation for more robust, context-aware model selection and analysis in multimodal AI. Future directions include extending PCRI to sequential, video, and audio domains, enabling further advances in real-world robustness.

\section{Limitations}

While PCRI provides a robust, interpretable signal of context sensitivity in MLLMs, several limitations remain. First, the current analysis is restricted to English-language vision-language datasets; extending PCRI to multilingual and cross-cultural tasks is an important next step. Second, PCRI measures robustness at the granularity of patches, but does not explicitly account for dependencies across patches or sequential/temporal context, future work could address these aspects. Third, our framework focuses on task-level context sensitivity, rather than image property variations such as resolution or noise; integrating these factors remains an open challenge.

Finally, while we validate PCRI alignment with human reasoning across two task types, broader studies, including more diverse datasets and annotator pools, would further strengthen the generalizability of our findings. We encourage the community to adopt and extend PCRI as a diagnostic tool for developing and deploying trustworthy, context-robust multimodal models.

% % Bibliography entries for the entire Anthology, followed by custom entries
% %\bibliography{anthology,custom}
% % Custom bibliography entries only
\bibliography{custom}

\begin{thebibliography}{48}
\providecommand{\natexlab}[1]{#1}

\bibitem[{Abdin et~al.(2024)Abdin, Aneja, Awadalla, Awadallah, Awan, Bach, Bahree, Bakhtiari, Bao, Behl et~al.}]{abdin2024phi}
Marah Abdin, Jyoti Aneja, Hany Awadalla, Ahmed Awadallah, Ammar~Ahmad Awan, Nguyen Bach, Amit Bahree, Arash Bakhtiari, Jianmin Bao, Harkirat Behl, et~al. 2024.
\newblock Phi-3 technical report: A highly capable language model locally on your phone.
\newblock \emph{arXiv preprint arXiv:2404.14219}.

\bibitem[{Agarwal et~al.(2024{\natexlab{a}})Agarwal, Pachauri, Zadeh, and Qian}]{agarwal2024techniques}
Amit Agarwal, Kulbhushan Pachauri, Iman Zadeh, and Jun Qian. 2024{\natexlab{a}}.
\newblock Techniques for graph data structure augmentation.
\newblock US Patent 11,989,964.

\bibitem[{Agarwal et~al.(2025{\natexlab{a}})Agarwal, Panda, Charles, Patel, Kumar, Pattnayak, Rafi, Kumar, Meghwani, Gupta, and Chae}]{agarwal-etal-2025-mvtamperbench}
Amit Agarwal, Srikant Panda, Angeline Charles, Hitesh~Laxmichand Patel, Bhargava Kumar, Priyaranjan Pattnayak, Taki~Hasan Rafi, Tejaswini Kumar, Hansa Meghwani, Karan Gupta, and Dong-Kyu Chae. 2025{\natexlab{a}}.
\newblock \href {https://doi.org/10.18653/v1/2025.findings-acl.963} {{MVT}amper{B}ench: Evaluating robustness of vision-language models}.
\newblock In \emph{Findings of the Association for Computational Linguistics: ACL 2025}, pages 18804--18828, Vienna, Austria. Association for Computational Linguistics.

\bibitem[{Agarwal et~al.(2024{\natexlab{b}})Agarwal, Panda, and Pachauri}]{agarwal2024synthetic}
Amit Agarwal, Srikant Panda, and Kulbhushan Pachauri. 2024{\natexlab{b}}.
\newblock Synthetic document generation pipeline for training artificial intelligence models.
\newblock US Patent App. 17/994,712.

\bibitem[{Agarwal et~al.(2025{\natexlab{b}})Agarwal, Panda, and Pachauri}]{agarwal-etal-2025-fs}
Amit Agarwal, Srikant Panda, and Kulbhushan Pachauri. 2025{\natexlab{b}}.
\newblock \href {https://aclanthology.org/2025.coling-industry.9/} {{FS}-{DAG}: Few shot domain adapting graph networks for visually rich document understanding}.
\newblock In \emph{Proceedings of the 31st International Conference on Computational Linguistics: Industry Track}, pages 100--114, Abu Dhabi, UAE. Association for Computational Linguistics.

\bibitem[{Agrawal et~al.(2016)Agrawal, Batra, and Parikh}]{agrawal2016analyzing}
Aishwarya Agrawal, Dhruv Batra, and Devi Parikh. 2016.
\newblock Analyzing the behavior of visual question answering models.
\newblock \emph{arXiv preprint arXiv:1606.07356}.

\bibitem[{Agrawal et~al.(2024)Agrawal, Antoniak, Hanna, Bout, Chaplot, Chudnovsky, Costa, Monicault, Garg, Gervet, Ghosh, Héliou, Jacob, Jiang, Khandelwal, Lacroix, Lample, Casas, Lavril, Scao, Lo, Marshall, Martin, Mensch, Muddireddy, Nemychnikova, Pellat, Platen, Raghuraman, Rozière, Sablayrolles, Saulnier, Sauvestre, Shang, Soletskyi, Stewart, Stock, Studnia, Subramanian, Vaze, Wang, and Yang}]{agrawal2024pixtral12b}
Pravesh Agrawal, Szymon Antoniak, Emma~Bou Hanna, Baptiste Bout, Devendra Chaplot, Jessica Chudnovsky, Diogo Costa, Baudouin~De Monicault, Saurabh Garg, Theophile Gervet, Soham Ghosh, Amélie Héliou, Paul Jacob, Albert~Q. Jiang, Kartik Khandelwal, Timothée Lacroix, Guillaume Lample, Diego~Las Casas, Thibaut Lavril, Teven~Le Scao, Andy Lo, William Marshall, Louis Martin, Arthur Mensch, Pavankumar Muddireddy, Valera Nemychnikova, Marie Pellat, Patrick~Von Platen, Nikhil Raghuraman, Baptiste Rozière, Alexandre Sablayrolles, Lucile Saulnier, Romain Sauvestre, Wendy Shang, Roman Soletskyi, Lawrence Stewart, Pierre Stock, Joachim Studnia, Sandeep Subramanian, Sagar Vaze, Thomas Wang, and Sophia Yang. 2024.
\newblock \href {https://arxiv.org/abs/2410.07073} {Pixtral 12b}.
\newblock \emph{Preprint}, arXiv:2410.07073.

\bibitem[{Chen et~al.(2024{\natexlab{a}})Chen, Ni, Huang, Liu, Jeong, Wen, Bastian, Latapie, and Imani}]{chen2024vltpvisionlanguageguidedtoken}
Hanning Chen, Yang Ni, Wenjun Huang, Yezi Liu, SungHeon Jeong, Fei Wen, Nathaniel Bastian, Hugo Latapie, and Mohsen Imani. 2024{\natexlab{a}}.
\newblock \href {https://arxiv.org/abs/2409.08464} {Vltp: Vision-language guided token pruning for task-oriented segmentation}.
\newblock \emph{Preprint}, arXiv:2409.08464.

\bibitem[{Chen et~al.(2024{\natexlab{b}})Chen, Li, Dong, Zhang, Zang, Chen, Duan, Wang, Qiao, Lin, and Zhao}]{chen2024rightwayevaluatinglarge}
Lin Chen, Jinsong Li, Xiaoyi Dong, Pan Zhang, Yuhang Zang, Zehui Chen, Haodong Duan, Jiaqi Wang, Yu~Qiao, Dahua Lin, and Feng Zhao. 2024{\natexlab{b}}.
\newblock \href {https://arxiv.org/abs/2403.20330} {Are we on the right way for evaluating large vision-language models?}
\newblock \emph{Preprint}, arXiv:2403.20330.

\bibitem[{Chen et~al.(2025)Chen, Wu, Liu, Pan, Liu, Xie, Yu, and Ruan}]{chen2025janus}
Xiaokang Chen, Zhiyu Wu, Xingchao Liu, Zizheng Pan, Wen Liu, Zhenda Xie, Xingkai Yu, and Chong Ruan. 2025.
\newblock Janus-pro: Unified multimodal understanding and generation with data and model scaling.
\newblock \emph{arXiv preprint arXiv:2501.17811}.

\bibitem[{Chen et~al.(2015)Chen, Fang, Lin, Vedantam, Gupta, Dollar, and Zitnick}]{chen2015microsoftcococaptionsdata}
Xinlei Chen, Hao Fang, Tsung-Yi Lin, Ramakrishna Vedantam, Saurabh Gupta, Piotr Dollar, and C.~Lawrence Zitnick. 2015.
\newblock \href {https://arxiv.org/abs/1504.00325} {Microsoft coco captions: Data collection and evaluation server}.
\newblock \emph{Preprint}, arXiv:1504.00325.

\bibitem[{Chen et~al.(2024{\natexlab{c}})Chen, Wang, Cao, Liu, Gao, Cui, Zhu, Ye, Tian, Liu et~al.}]{chen2024expanding}
Zhe Chen, Weiyun Wang, Yue Cao, Yangzhou Liu, Zhangwei Gao, Erfei Cui, Jinguo Zhu, Shenglong Ye, Hao Tian, Zhaoyang Liu, et~al. 2024{\natexlab{c}}.
\newblock Expanding performance boundaries of open-source multimodal models with model, data, and test-time scaling.
\newblock \emph{arXiv preprint arXiv:2412.05271}.

\bibitem[{Chen et~al.(2024{\natexlab{d}})Chen, Wang, Tian, Ye, Gao, Cui, Tong, Hu, Luo, Ma et~al.}]{chen2024far}
Zhe Chen, Weiyun Wang, Hao Tian, Shenglong Ye, Zhangwei Gao, Erfei Cui, Wenwen Tong, Kongzhi Hu, Jiapeng Luo, Zheng Ma, et~al. 2024{\natexlab{d}}.
\newblock How far are we to gpt-4v? closing the gap to commercial multimodal models with open-source suites.
\newblock \emph{arXiv preprint arXiv:2404.16821}.

\bibitem[{Chen et~al.(2024{\natexlab{e}})Chen, Wu, Wang, Su, Chen, Xing, Zhong, Zhang, Zhu, Lu et~al.}]{chen2024internvl}
Zhe Chen, Jiannan Wu, Wenhai Wang, Weijie Su, Guo Chen, Sen Xing, Muyan Zhong, Qinglong Zhang, Xizhou Zhu, Lewei Lu, et~al. 2024{\natexlab{e}}.
\newblock Internvl: Scaling up vision foundation models and aligning for generic visual-linguistic tasks.
\newblock In \emph{Proceedings of the IEEE/CVF Conference on Computer Vision and Pattern Recognition}, pages 24185--24198.

\bibitem[{Deitke et~al.(2024)Deitke, Clark, Lee, Tripathi, Yang, Park, Salehi, Muennighoff, Lo, Soldaini, Lu, Anderson, Bransom, Ehsani, Ngo, Chen, Patel, Yatskar, Callison-Burch, Head, Hendrix, Bastani, VanderBilt, Lambert, Chou, Chheda, Sparks, Skjonsberg, Schmitz, Sarnat, Bischoff, Walsh, Newell, Wolters, Gupta, Zeng, Borchardt, Groeneveld, Nam, Lebrecht, Wittlif, Schoenick, Michel, Krishna, Weihs, Smith, Hajishirzi, Girshick, Farhadi, and Kembhavi}]{deitke2024molmopixmoopenweights}
Matt Deitke, Christopher Clark, Sangho Lee, Rohun Tripathi, Yue Yang, Jae~Sung Park, Mohammadreza Salehi, Niklas Muennighoff, Kyle Lo, Luca Soldaini, Jiasen Lu, Taira Anderson, Erin Bransom, Kiana Ehsani, Huong Ngo, YenSung Chen, Ajay Patel, Mark Yatskar, Chris Callison-Burch, Andrew Head, Rose Hendrix, Favyen Bastani, Eli VanderBilt, Nathan Lambert, Yvonne Chou, Arnavi Chheda, Jenna Sparks, Sam Skjonsberg, Michael Schmitz, Aaron Sarnat, Byron Bischoff, Pete Walsh, Chris Newell, Piper Wolters, Tanmay Gupta, Kuo-Hao Zeng, Jon Borchardt, Dirk Groeneveld, Crystal Nam, Sophie Lebrecht, Caitlin Wittlif, Carissa Schoenick, Oscar Michel, Ranjay Krishna, Luca Weihs, Noah~A. Smith, Hannaneh Hajishirzi, Ross Girshick, Ali Farhadi, and Aniruddha Kembhavi. 2024.
\newblock \href {https://arxiv.org/abs/2409.17146} {Molmo and pixmo: Open weights and open data for state-of-the-art vision-language models}.
\newblock \emph{Preprint}, arXiv:2409.17146.

\bibitem[{Duan et~al.(2024)Duan, Yang, Qiao, Fang, Chen, Liu, Dong, Zang, Zhang, Wang et~al.}]{duan2024vlmevalkit}
Haodong Duan, Junming Yang, Yuxuan Qiao, Xinyu Fang, Lin Chen, Yuan Liu, Xiaoyi Dong, Yuhang Zang, Pan Zhang, Jiaqi Wang, et~al. 2024.
\newblock Vlmevalkit: An open-source toolkit for evaluating large multi-modality models.
\newblock In \emph{Proceedings of the 32nd ACM international conference on multimedia}, pages 11198--11201.

\bibitem[{Fu et~al.(2024{\natexlab{a}})Fu, Chen, Shen, Qin, Zhang, Lin, Yang, Zheng, Li, Sun, Wu, and Ji}]{fu2024mmecomprehensiveevaluationbenchmark}
Chaoyou Fu, Peixian Chen, Yunhang Shen, Yulei Qin, Mengdan Zhang, Xu~Lin, Jinrui Yang, Xiawu Zheng, Ke~Li, Xing Sun, Yunsheng Wu, and Rongrong Ji. 2024{\natexlab{a}}.
\newblock \href {https://arxiv.org/abs/2306.13394} {Mme: A comprehensive evaluation benchmark for multimodal large language models}.
\newblock \emph{Preprint}, arXiv:2306.13394.

\bibitem[{Fu et~al.(2024{\natexlab{b}})Fu, Hu, Li, Feng, Wang, Lin, Roth, Smith, Ma, and Krishna}]{fu2024blinkmultimodallargelanguage}
Xingyu Fu, Yushi Hu, Bangzheng Li, Yu~Feng, Haoyu Wang, Xudong Lin, Dan Roth, Noah~A. Smith, Wei-Chiu Ma, and Ranjay Krishna. 2024{\natexlab{b}}.
\newblock \href {https://arxiv.org/abs/2404.12390} {Blink: Multimodal large language models can see but not perceive}.
\newblock \emph{Preprint}, arXiv:2404.12390.

\bibitem[{Gao et~al.(2024)Gao, Chen, Cui, Ren, Wang, Zhu, Tian, Ye, He, Zhu et~al.}]{gao2024mini}
Zhangwei Gao, Zhe Chen, Erfei Cui, Yiming Ren, Weiyun Wang, Jinguo Zhu, Hao Tian, Shenglong Ye, Junjun He, Xizhou Zhu, et~al. 2024.
\newblock Mini-internvl: A flexible-transfer pocket multimodal model with 5\% parameters and 90\% performance.
\newblock \emph{arXiv preprint arXiv:2410.16261}.

\bibitem[{Guan et~al.(2024)Guan, Liu, Wu, Xian, Li, Liu, Wang, Chen, Huang, Yacoob, Manocha, and Zhou}]{guan2024hallusionbenchadvanceddiagnosticsuite}
Tianrui Guan, Fuxiao Liu, Xiyang Wu, Ruiqi Xian, Zongxia Li, Xiaoyu Liu, Xijun Wang, Lichang Chen, Furong Huang, Yaser Yacoob, Dinesh Manocha, and Tianyi Zhou. 2024.
\newblock \href {https://arxiv.org/abs/2310.14566} {Hallusionbench: An advanced diagnostic suite for entangled language hallucination and visual illusion in large vision-language models}.
\newblock \emph{Preprint}, arXiv:2310.14566.

\bibitem[{Gurari et~al.(2018)Gurari, Li, Stangl, Guo, Lin, Grauman, Luo, and Bigham}]{gurari2018vizwizgrandchallengeanswering}
Danna Gurari, Qing Li, Abigale~J. Stangl, Anhong Guo, Chi Lin, Kristen Grauman, Jiebo Luo, and Jeffrey~P. Bigham. 2018.
\newblock \href {https://arxiv.org/abs/1802.08218} {Vizwiz grand challenge: Answering visual questions from blind people}.
\newblock \emph{Preprint}, arXiv:1802.08218.

\bibitem[{Jiang et~al.(2022)Jiang, Xu, Li, Yan, Ye, Zhang, Bi, and Huang}]{jiang2022trips}
Chaoya Jiang, Haiyang Xu, Chenliang Li, Ming Yan, Wei Ye, Shikun Zhang, Bin Bi, and Songfang Huang. 2022.
\newblock Trips: Efficient vision-and-language pre-training with text-relevant image patch selection.
\newblock In \emph{Proceedings of the 2022 Conference on Empirical Methods in Natural Language Processing}, pages 4084--4096.

\bibitem[{Kembhavi et~al.(2016)Kembhavi, Salvato, Kolve, Seo, Hajishirzi, and Farhadi}]{kembhavi2016diagramworthdozenimages}
Aniruddha Kembhavi, Mike Salvato, Eric Kolve, Minjoon Seo, Hannaneh Hajishirzi, and Ali Farhadi. 2016.
\newblock \href {https://arxiv.org/abs/1603.07396} {A diagram is worth a dozen images}.
\newblock \emph{Preprint}, arXiv:1603.07396.

\bibitem[{Li et~al.(2024)Li, Zhang, Guo, Zhang, Li, Zhang, Zhang, Zhang, Li, Liu, and Li}]{li2024llavaonevisioneasyvisualtask}
Bo~Li, Yuanhan Zhang, Dong Guo, Renrui Zhang, Feng Li, Hao Zhang, Kaichen Zhang, Peiyuan Zhang, Yanwei Li, Ziwei Liu, and Chunyuan Li. 2024.
\newblock \href {https://arxiv.org/abs/2408.03326} {Llava-onevision: Easy visual task transfer}.
\newblock \emph{Preprint}, arXiv:2408.03326.

\bibitem[{Li et~al.(2025)Li, Zhang, Zhang, Wu, Tian, Sun, Lu, Min, Liu, Lin et~al.}]{li2025r}
Chunyi Li, Jianbo Zhang, Zicheng Zhang, Haoning Wu, Yuan Tian, Wei Sun, Guo Lu, Xiongkuo Min, Xiaohong Liu, Weisi Lin, et~al. 2025.
\newblock R-bench: Are your large multimodal model robust to real-world corruptions?
\newblock \emph{IEEE Journal of Selected Topics in Signal Processing}.

\bibitem[{Li et~al.(2023)Li, Du, Zhou, Wang, Zhao, and Wen}]{li2023evaluatingobjecthallucinationlarge}
Yifan Li, Yifan Du, Kun Zhou, Jinpeng Wang, Wayne~Xin Zhao, and Ji-Rong Wen. 2023.
\newblock \href {https://arxiv.org/abs/2305.10355} {Evaluating object hallucination in large vision-language models}.
\newblock \emph{Preprint}, arXiv:2305.10355.

\bibitem[{Lu et~al.(2022)Lu, Mishra, Xia, Qiu, Chang, Zhu, Tafjord, Clark, and Kalyan}]{lu2022learnexplainmultimodalreasoning}
Pan Lu, Swaroop Mishra, Tony Xia, Liang Qiu, Kai-Wei Chang, Song-Chun Zhu, Oyvind Tafjord, Peter Clark, and Ashwin Kalyan. 2022.
\newblock \href {https://arxiv.org/abs/2209.09513} {Learn to explain: Multimodal reasoning via thought chains for science question answering}.
\newblock \emph{Preprint}, arXiv:2209.09513.

\bibitem[{Lu et~al.(2024)Lu, Li, Chen, Xu, Luo, Zhang, and Ye}]{lu2024ovis}
Shiyin Lu, Yang Li, Qing-Guo Chen, Zhao Xu, Weihua Luo, Kaifu Zhang, and Han-Jia Ye. 2024.
\newblock Ovis: Structural embedding alignment for multimodal large language model.
\newblock \emph{arXiv preprint arXiv:2405.20797}.

\bibitem[{Ma et~al.(2024)Ma, Wang, Sun, Lin, Zhou, Ji, and Ji}]{ma2024infllava}
Yiwei Ma, Zhibin Wang, Xiaoshuai Sun, Weihuang Lin, Qiang Zhou, Jiayi Ji, and Rongrong Ji. 2024.
\newblock Inf-llava: Dual-perspective perception for high-resolution multimodal large language model.
\newblock \emph{arXiv preprint arXiv:2407.16198}.

\bibitem[{Masry et~al.(2022)Masry, Long, Tan, Joty, and Hoque}]{masry2022chartqabenchmarkquestionanswering}
Ahmed Masry, Do~Xuan Long, Jia~Qing Tan, Shafiq Joty, and Enamul Hoque. 2022.
\newblock \href {https://arxiv.org/abs/2203.10244} {Chartqa: A benchmark for question answering about charts with visual and logical reasoning}.
\newblock \emph{Preprint}, arXiv:2203.10244.

\bibitem[{Meghwani et~al.(2025)Meghwani, Agarwal, Pattnayak, Patel, and Panda}]{meghwani-etal-2025-hard}
Hansa Meghwani, Amit Agarwal, Priyaranjan Pattnayak, Hitesh~Laxmichand Patel, and Srikant Panda. 2025.
\newblock \href {https://doi.org/10.18653/v1/2025.acl-industry.72} {Hard negative mining for domain-specific retrieval in enterprise systems}.
\newblock In \emph{Proceedings of the 63rd Annual Meeting of the Association for Computational Linguistics (Volume 6: Industry Track)}, pages 1013--1026, Vienna, Austria. Association for Computational Linguistics.

\bibitem[{Panda et~al.(2025{\natexlab{a}})Panda, Hari, Panda, Agarwal, and Patel}]{panda2025whosaskinginvestigatingbias}
Srikant Panda, Vishnu Hari, Kalpana Panda, Amit Agarwal, and Hitesh~Laxmichand Patel. 2025{\natexlab{a}}.
\newblock \href {https://arxiv.org/abs/2508.15831} {Who's asking? investigating bias through the lens of disability framed queries in llms}.
\newblock \emph{Preprint}, arXiv:2508.15831.

\bibitem[{Panda et~al.(2025{\natexlab{b}})Panda, Patel, Al-Khalifa, Agarwal, Al-Khalifa, and Al-Ghamdi}]{panda2025daiqauditingdemographicattribute}
Srikant Panda, Hitesh~Laxmichand Patel, Shahad Al-Khalifa, Amit Agarwal, Hend Al-Khalifa, and Sharefah Al-Ghamdi. 2025{\natexlab{b}}.
\newblock \href {https://arxiv.org/abs/2508.15830} {Daiq: Auditing demographic attribute inference from question in llms}.
\newblock \emph{Preprint}, arXiv:2508.15830.

\bibitem[{Patel et~al.(2025)Patel, Agarwal, Das, Kumar, Panda, Pattnayak, Rafi, Kumar, and Chae}]{patel2025sweeval}
Hitesh~Laxmichand Patel, Amit Agarwal, Arion Das, Bhargava Kumar, Srikant Panda, Priyaranjan Pattnayak, Taki~Hasan Rafi, Tejaswini Kumar, and Dong-Kyu Chae. 2025.
\newblock Sweeval: Do llms really swear? a safety benchmark for testing limits for enterprise use.
\newblock In \emph{Proceedings of the 2025 Conference of the Nations of the Americas Chapter of the Association for Computational Linguistics: Human Language Technologies (Volume 3: Industry Track)}, pages 558--582.

\bibitem[{Patel et~al.(2024)Patel, Agarwal, Kumar, Gupta, and Pattnayak}]{patel2024llm}
Hitesh~Laxmichand Patel, Amit Agarwal, Bhargava Kumar, Karan Gupta, and Priyaranjan Pattnayak. 2024.
\newblock Llm for barcodes: Generating diverse synthetic data for identity documents.
\newblock \emph{arXiv preprint arXiv:2411.14962}.

\bibitem[{Pattnayak et~al.(2025)Pattnayak, Agarwal, Meghwani, Patel, and Panda}]{pattnayak2025hybrid}
Priyaranjan Pattnayak, Amit Agarwal, Hansa Meghwani, Hitesh~Laxmichand Patel, and Srikant Panda. 2025.
\newblock Hybrid ai for responsive multi-turn online conversations with novel dynamic routing and feedback adaptation.
\newblock In \emph{Proceedings of the 4th International Workshop on Knowledge-Augmented Methods for Natural Language Processing}, pages 215--229.

\bibitem[{Pattnayak et~al.(2024)Pattnayak, Patel, Kumar, Agarwal, Banerjee, Panda, and Kumar}]{pattnayak2024survey}
Priyaranjan Pattnayak, Hitesh~Laxmichand Patel, Bhargava Kumar, Amit Agarwal, Ishan Banerjee, Srikant Panda, and Tejaswini Kumar. 2024.
\newblock Survey of large multimodal model datasets, application categories and taxonomy.
\newblock \emph{arXiv preprint arXiv:2412.17759}.

\bibitem[{Qiu et~al.(2024)Qiu, Zhu, Shi, Wenzel, Tang, Zhao, Li, and Li}]{qiu2024benchmarkingrobustnessmultimodalimagetext}
Jielin Qiu, Yi~Zhu, Xingjian Shi, Florian Wenzel, Zhiqiang Tang, Ding Zhao, Bo~Li, and Mu~Li. 2024.
\newblock \href {https://arxiv.org/abs/2212.08044} {Benchmarking robustness of multimodal image-text models under distribution shift}.
\newblock \emph{Preprint}, arXiv:2212.08044.

\bibitem[{Singh et~al.(2019)Singh, Natarajan, Shah, Jiang, Chen, Batra, Parikh, and Rohrbach}]{singh2019vqamodelsread}
Amanpreet Singh, Vivek Natarajan, Meet Shah, Yu~Jiang, Xinlei Chen, Dhruv Batra, Devi Parikh, and Marcus Rohrbach. 2019.
\newblock \href {https://arxiv.org/abs/1904.08920} {Towards vqa models that can read}.
\newblock \emph{Preprint}, arXiv:1904.08920.

\bibitem[{Thapa et~al.(2024)Thapa, Chen, Covert, Chalamala, Athiwaratkun, Song, and Zou}]{thapa2024dragonfly}
Rahul Thapa, Kezhen Chen, Ian Covert, Rahul Chalamala, Ben Athiwaratkun, Shuaiwen~Leon Song, and James Zou. 2024.
\newblock Dragonfly: Multi-resolution zoom-in encoding enhances vision-language models.
\newblock \emph{arXiv preprint arXiv:2406.00977}.

\bibitem[{Wang et~al.(2024{\natexlab{a}})Wang, Wang, Xu, Zhang, Gu, Jia, Wang, Xu, Yan, Zhang, and Sang}]{wang2024amberllmfreemultidimensionalbenchmark}
Junyang Wang, Yuhang Wang, Guohai Xu, Jing Zhang, Yukai Gu, Haitao Jia, Jiaqi Wang, Haiyang Xu, Ming Yan, Ji~Zhang, and Jitao Sang. 2024{\natexlab{a}}.
\newblock \href {https://arxiv.org/abs/2311.07397} {Amber: An llm-free multi-dimensional benchmark for mllms hallucination evaluation}.
\newblock \emph{Preprint}, arXiv:2311.07397.

\bibitem[{Wang et~al.(2024{\natexlab{b}})Wang, Bai, Tan, Wang, Fan, Bai, Chen, Liu, Wang, Ge, Fan, Dang, Du, Ren, Men, Liu, Zhou, Zhou, and Lin}]{wang2024qwen2}
Peng Wang, Shuai Bai, Sinan Tan, Shijie Wang, Zhihao Fan, Jinze Bai, Keqin Chen, Xuejing Liu, Jialin Wang, Wenbin Ge, Yang Fan, Kai Dang, Mengfei Du, Xuancheng Ren, Rui Men, Dayiheng Liu, Chang Zhou, Jingren Zhou, and Junyang Lin. 2024{\natexlab{b}}.
\newblock \href {https://arxiv.org/abs/2409.12191} {Qwen2-vl: Enhancing vision-language model's perception of the world at any resolution}.
\newblock \emph{Preprint}, arXiv:2409.12191.

\bibitem[{XAI-Org(2024)}]{RealWorldQA}
XAI-Org. 2024.
\newblock Realworldqa dataset.
\newblock \url{https://huggingface.co/datasets/xai-org/RealworldQA}.
\newblock Accessed: 2024-03-15.

\bibitem[{Ye et~al.(2024)Ye, Jiang, Xu, Ye, Li, Yan, Zhang, Huang, and Huang}]{ye2024efficient}
Wei Ye, Chaoya Jiang, Haiyang Xu, Chenhao Ye, Chenliang Li, Ming Yan, Shikun Zhang, Songhang Huang, and Fei Huang. 2024.
\newblock Efficient vision-and-language pre-training with text-relevant image patch selection.
\newblock \emph{arXiv preprint arXiv:2403.07883}.

\bibitem[{Zhang et~al.(2024)Zhang, Hu, Khayatkhoei, Ilievski, and Sun}]{zhang2024exploringperceptuallimitationmultimodal}
Jiarui Zhang, Jinyi Hu, Mahyar Khayatkhoei, Filip Ilievski, and Maosong Sun. 2024.
\newblock \href {https://arxiv.org/abs/2402.07384} {Exploring perceptual limitation of multimodal large language models}.
\newblock \emph{Preprint}, arXiv:2402.07384.

\bibitem[{Zhang et~al.(2023)Zhang, Khayatkhoei, Chhikara, and Ilievski}]{zhang2023small}
Jiarui Zhang, Mahyar Khayatkhoei, Prateek Chhikara, and Filip Ilievski. 2023.
\newblock Towards perceiving small visual details in zero-shot visual question answering with multimodal llms.
\newblock \emph{arXiv preprint arXiv:2310.16033}.

\bibitem[{Zhang et~al.(2025)Zhang, Wang, Li, Zhang, Taslakian, Rajeswar, Fu, Liu, and Bengio}]{zhang2025vcr}
Tianyu Zhang, Suyuchen Wang, Lu~Li, Ge~Zhang, Perouz Taslakian, Sai Rajeswar, Jie Fu, Bang Liu, and Yoshua Bengio. 2025.
\newblock Vcr: A task for pixel-level complex reasoning in vision language models via restoring occluded text.
\newblock In \emph{The Thirteenth International Conference on Learning Representations}.

\bibitem[{Zhu et~al.(2024)Zhu, Dong, Song, Guo, and Zheng}]{zhu2024hyvilm}
Shiding Zhu, Wenhui Dong, Jun Song, Yanan Guo, and Bo~Zheng. 2024.
\newblock Hyvilm: Enhancing fine-grained recognition with a hybrid encoder for vision-language models.
\newblock \emph{arXiv preprint arXiv:2412.08378}.

\end{thebibliography}

\clearpage

\appendix

\section{Appendix}
\label{sec:appendix}

\subsection{Detailed Methodology: PCRI }
\label{sec:PCRI_properties}

Let $\mathcal{D} = \{(x^{(i)}, q^{(i)}, a^{(i)})\}_{i=1}^N$ denote a dataset of $N$ image–query–answer triples, where $x^{(i)}$ is an image, $q^{(i)}$ is the associated query (e.g., question or prompt), and $a^{(i)}$ is the ground-truth answer.

Let $M(\cdot)$ be the evaluation metric for the task (e.g., accuracy, BLEU, F1), computed over $\mathcal{D}$ as per the standard benchmark protocol.

The model’s performance on the dataset using the full image is:
\begin{equation}
    P_{\text{whole}} = M\left(\{(x^{(i)}, q^{(i)}, a^{(i)})\}_{i=1}^N\right).
    \label{eq:pwhole}
\end{equation}

For each image $x^{(i)}$, let $\{x^{(i, j)}\}_{j=1}^{n^2}$ denote its $n \times n$ non-overlapping patches. The patch-based performance at granularity $n$ is:
\begin{equation}
    P_{\text{patch}, n} = M\left(\{(x^{(i, j^*)}, q^{(i)}, a^{(i)})\}_{i=1}^N\right),
    \label{eq:ppatch}
\end{equation}
where $j^* = \arg\max_{j} s^{(i, j)}$ is the index of the patch with the highest model performance $s^{(i, j)}$ for instance $i$ (with $s^{(i, j)}$ computed per the metric $M$).

The \textbf{Patch Context Robustness Index (PCRI)} at granularity $n$ is:
\begin{equation}
    \text{PCRI}_n = 1 - \frac{P_{\text{patch}, n}}{P_{\text{whole}}}.
    \label{eq:pcri}
\end{equation}

PCRI thus quantifies the (relative) performance drop or gain when the model is restricted to its best-performing local patch versus the full image context. Because $P_{\text{whole}}$ and $P_{\text{patch}, n}$ are computed with the same metric $M$ over the same dataset, PCRI is \textit{invariant} to the metric’s scale and is directly comparable across tasks and datasets.

\paragraph{Interpretation.}
\begin{itemize}
    \item $\text{PCRI}_n \approx 0$: Model is robust—global and local context yield similar performance.
    \item $\text{PCRI}_n > 0$: Model requires global context; patch-only input reduces performance.
    \item $\text{PCRI}_n < 0$: Model is distracted by global context or benefits from local-only cues.
\end{itemize}
PCRI is undefined if $P_{\text{whole}} = 0$ for a given model/task pair, as division by zero is not meaningful. In such cases, the model cannot solve the task even with full context.

\paragraph{Rationale for Max Aggregation.}
For each instance, we select the patch $j^*$ with the highest model performance $s^{(i, j)}$. This “max” aggregation captures whether the model can solve the task using \emph{any} local patch. This is robust for both discrete and continuous metrics and highlights cases where a model is brittle to added context or reliant on specific local evidence. In contrast, mean or sum aggregation could mask context brittleness by averaging over patches that may be trivially correct or uninformative. The max operation thus yields a clearer, more actionable signal for model selection and robustness analysis.

\paragraph{Granularity sensitivity and compute.}
\label{sec:appendix_granular}
We ablate $n\!\in\!\{3,4,5\}$ on selected model and tasks (Table~\ref{tab:pcri_granularity}). Increasing $n$ (finer granularity) can expose local solvability but increases evaluation cost quadratically ($n^2$ patch forward passes per image). For $n{>}3$, patches often become too small to capture meaningful semantics; PCRI’s discriminative power saturates and may drop because the task becomes infeasible under heavy fragmentation. In practice, $n\!=\!2$ or $n\!=\!3$ provides a strong trade-off between informativeness and efficiency.

\begin{table}[!ht]
    \centering
    \scalebox{0.85}{
    \renewcommand{\arraystretch}{1.1}
    \begin{tabular}{|l|c|c|c|}
        \hline
        \textbf{Dataset} & \textbf{PCRI$_{n=3}$} & \textbf{PCRI$_{n=4}$} & \textbf{PCRI$_{n=5}$} \\
        \hline
        ChartQA\_TEST & 0.237 & 0.300 & 1.000 \\
        AMBER         & -0.038 & -0.030 & 0.550 \\
        BLINK         & -0.516 & -0.558 & 0.380 \\
        \hline
    \end{tabular}}
    \caption{PCRI values (Molmo-1B) at increasing patch granularity ($n{=}3,4,5$) on representative benchmarks. More negative $\Rightarrow$ stronger global-context distraction (best patch $\ge$ full image). Very large positive values at $n{=}5$ indicate fragmentation-induced unsolvability rather than new trends, motivating our default of $n\!\in\!\{2,3\}$.}
    \label{tab:pcri_granularity}
\end{table}

\subsection{Chance floors and Validity gate}
\label{sec:chance-validity}
\paragraph{Chance floors.}
For each dataset $d$, we define $C(d)$ as the task’s random baseline: $1/|\mathcal{Y}|$ for balanced $|\mathcal{Y}|$-way classification; the class-prior baseline for imbalanced classification; and the documented random/shuffle baseline for retrieval/captioning metrics. When an official baseline is unavailable, we estimate $C(d)$ via random-shuffle following standard protocol.

\paragraph{How to set $C(d)$.}
\textit{Classification:} $C(d)=1/|\mathcal{Y}|$ if balanced; otherwise use the empirical class-prior baseline.
\textit{Retrieval:} for $N$ candidates and one relevant item, $C_{\text{R@}K}\!\approx\!K/N$; if multiple relevants or nonstandard pools, use the dataset’s documented random baseline or Monte Carlo shuffle.
\textit{Captioning/QA metrics:} use the documented random or shuffle baseline provided by dataset authors.

\paragraph{Uncertainty and gate application.}
We estimate $\mathrm{SE}$ for $P_{\text{whole}}$ via nonparametric bootstrap over evaluation examples ($B{=}1000$ by default) and apply the gate $P_{\text{whole}} \ge C(d) + \max\{\delta, 2\,\mathrm{SE}\}$ with $\delta{=}0.01$ on $[0,1]$-scaled metrics.

\paragraph{Reporting policy.}
If a model–dataset pair fails the gate, do \emph{not} compute/interpret PCRI; report only the underlying task metrics and mark PCRI as N/A. In our experiments, all model–dataset pairs satisfy the gate; we therefore interpret PCRI for all results.

% \paragraph{Applicability in this study.}
% All model–dataset pairs satisfy the gate; hence PCRI is reported and interpreted throughout.

% \section{Chance floors and validity gate}\label{app:chance-validity}
% \paragraph{How to set $C(d)$.}
% \textit{Classification:} $C(d)=1/|\mathcal{Y}|$ if balanced; otherwise use the empirical class-prior baseline.
% \textit{Retrieval:} for $N$ candidates and one relevant item, $C_{\text{R@}K}\!\approx\!K/N$; if multiple relevants or nonstandard pools, use the dataset’s documented random baseline or Monte Carlo shuffle.
% \textit{Captioning/QA metrics:} use the documented random or shuffle baseline provided by dataset authors.

% \paragraph{Uncertainty.}
% Compute $\mathrm{SE}$ of $P_{\text{whole}}$ via nonparametric bootstrap over examples ($B{=}1000$). Apply the gate $P_{\text{whole}} \ge C(d) + \max\{\delta, 2\,\mathrm{SE}\}$ with $\delta{=}0.01$ on $[0,1]$-scaled metrics.

% \paragraph{Reporting policy.}
% If a model–dataset pair fails the gate, do \emph{not} compute/interpret PCRI; report only the underlying task metrics and mark PCRI as N/A. In our experiments, all model–dataset pairs satisfy the gate; we therefore interpret PCRI for all results.

\subsection{Benchmarks \& Models}
\label{sec:data_models}
We evaluate our approach across 15 widely-used vision-language benchmarks, covering a diverse range of tasks. These datasets were selected to represent a balanced mix of localized perception tasks (e.g. object recognition) and global contextual reasoning challenges (e.g., complex multi- choice question answering).

\begin{table*}[!ht]
\centering

\scalebox{0.95}{
\begin{tabular}{lcccc}
\toprule
Dataset & PCRI & Full Img. Perf. & Patch Perf. & $\Delta$ Perf. \\
\midrule
AI2D & -0.23 & 96.7\% & 96.7\% & 0\% \\
RealWorldQA & -0.29 & 98\% & 96\% & -2\% \\
\bottomrule
\end{tabular}
}
\caption{Human study accuracy (\%) in patch vs. full image conditions, compared to model PCRI(InternVL-2.5-1B) scores.}
\label{tab:human_study_pcri}
\end{table*}

\begin{figure*}[!ht]
    \centering
    \includegraphics[width=\linewidth]{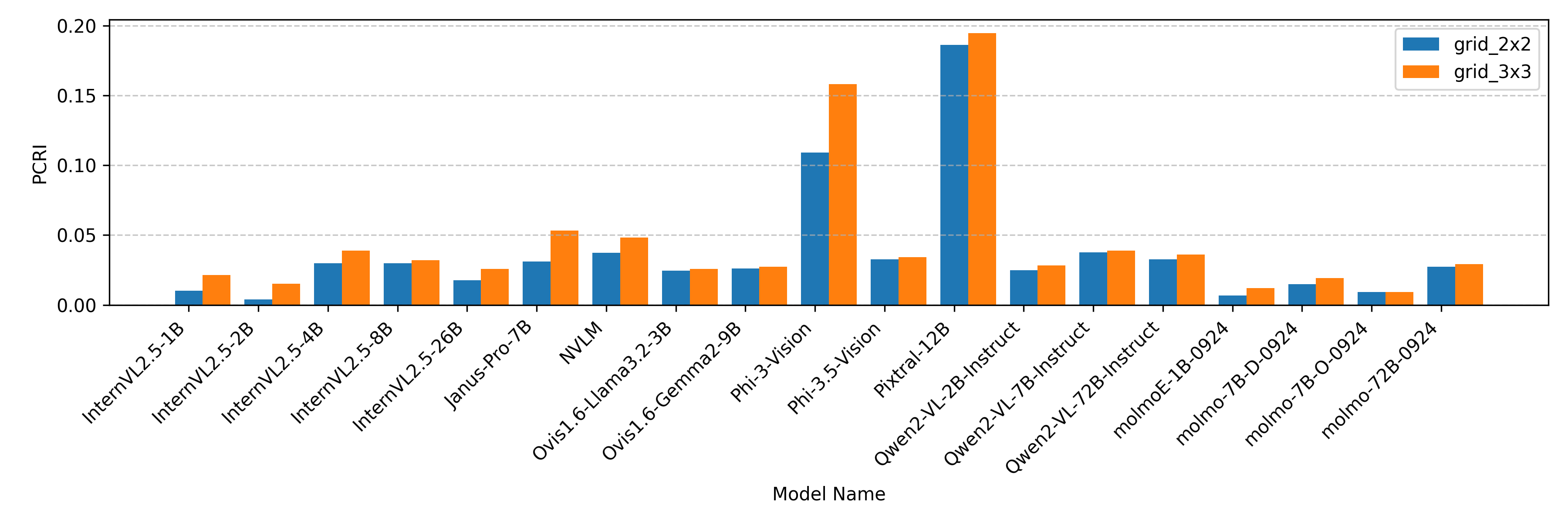}
    \caption{Avg. PCRI scores across different MLLMs for VQA tasks at patch granularities $2\times2$ and $3\times3$  Positive PCRI values indicate improved model performance with full-image contexts.%, whereas negative values indicate superior performance on localized patches.
    }
    \vspace{-1em}
    \label{fig:vqa_avg}
\end{figure*}

Our selection employs diverse realworld datasets that inherently contain a wide range of variations in image resolution, background complexity, and visual distortions. Consequently, PCRI metric has been rigorously tested across these naturally occurring variations, providing strong evidence of its robustness and practical relevance under realistic, heterogeneous visual conditions. Future work could complement these findings with controlled ablation studies to isolate the impact of each factor.

\begin{itemize}
    \item \textbf{Visual Question Answering (VQA):}  
    Benchmarks such as GQA \cite{lu2022learnexplainmultimodalreasoning}, ChartQA \cite{masry2022chartqabenchmarkquestionanswering}, TextVQA \cite{singh2019vqamodelsread}, and VizWiz \cite{gurari2018vizwizgrandchallengeanswering} are open-ended VQA tasks where MLLMs must generate responses without restricted answer choices. These benchmarks assess a model's ability to infer answers based on both localized and global scene information.

    \item \textbf{Multiple-Choice Question Answering (MCQ):}  
    Benchmarks including BLINK \cite{fu2024blinkmultimodallargelanguage}, RealWorldQA \cite{RealWorldQA}, AI2D \cite{kembhavi2016diagramworthdozenimages}, ScienceQA \cite{lu2022learnexplainmultimodalreasoning}, and MMStar \cite{chen2024rightwayevaluatinglarge} provide multiple answer choices, requiring MLLMs to differentiate between options and select the most accurate response. These tasks evaluate multimodal reasoning and answer disambiguation, offering insights into whether datasets support global contextual reasoning.

    \item \textbf{Yes/No Questions (Binary Classification):}  
    Datasets such as POPE \cite{li2023evaluatingobjecthallucinationlarge}, HallusionBench \cite{guan2024hallusionbenchadvanceddiagnosticsuite}, AMBER \cite{wang2024amberllmfreemultidimensionalbenchmark}, and MME \cite{fu2024mmecomprehensiveevaluationbenchmark} focus on binary (yes/no) questions. AMBER, in particular, tests the ability of the model to capture fine-grained spatial relationships, making it useful for evaluating whether a dataset enforces strict spatial comprehension.

    \item \textbf{Image Captioning and Semantic Understanding:}  
    We include MS-COCO (COCO Captions) \cite{chen2015microsoftcococaptionsdata} to evaluate semantic understanding at the global level. Captioning tasks assess whether models can generate accurate textual descriptions based on an entire image rather than relying on isolated object-level information.
\end{itemize}

\paragraph{Models:}  
We benchmark 19 state-of-the-art MLLMs, including InternVL \cite{chen2024far,gao2024mini,chen2024expanding,chen2024internvl}, Janu, LlaVaOneVision \cite{li2024llavaonevisioneasyvisualtask}, Molmo \cite{deitke2024molmopixmoopenweights}, Qwen2-VL \cite{wang2024qwen2}, Phi-3 Series \cite{abdin2024phi}, Ovis \cite{lu2024ovis} and Pixtral \cite{agrawal2024pixtral12b} across various model sizes.

\subsection{Extended Results}
\label{sec:extended_results}

This section provides detailed experimental results to supplement the main paper, including human-study, model and task-level PCRI breakdowns, granular ablations, and additional qualitative analyses. These extended results offer further evidence for the robustness patterns and context sensitivity described in the main text, supporting both reproducibility and deeper diagnostic insight for practitioners and researchers.

% \noindent Each evaluation is executed three times to account for inherent stochasticity in model predictions and input processing, such as random initialization and sampling variations. The averaged results exhibit negligible variance across runs, confirming the stability of our evaluation protocol. 

\subsubsection{Human Study: Protocol and Production Validation}
\label{sec:human_study_appendix}

\paragraph{Protocol.}
To assess whether PCRI aligns with human context reasoning, we conducted a controlled study on two representative benchmarks: AI2D and RealWorldQA. For each, we randomly sampled approximately 20\% of the data (AI2D: 300 samples; RealWorldQA: 80 samples). Three annotators (authors, blinded to model outputs and labels) independently answered each query in two settings: (1) with all the patches (one at a time), and (2) with the full image. Annotators strictly followed official task instructions and provided both answers and qualitative feedback for each condition.

\paragraph{Findings.}
Table~\ref{tab:human_study_pcri} summarizes results. In all cases, human accuracy with the full image was equal to or higher than with any patch: for AI2D, performance was nearly identical in both settings; for RealWorldQA, accuracy increased slightly with the full image (from 96\% patch-only to 98\%). Annotators consistently reported higher confidence and less ambiguity when presented with the full image, reinforcing the value of global context for human reasoning.

By contrast, several MLLMs exhibited \textbf{negative PCRI} on these benchmarks, performing better on patches than on the full image, behavior never observed in humans. This suggests that such models are either overfitting to spurious cues in isolated patches or are distracted by irrelevant global information, leading to brittle and non-human-like reasoning. Qualitative annotator feedback further supports this diagnosis, with patch-only settings described as “ambiguous” or “lacking key context.”

\paragraph{Production Deployment and User Feedback.}
To validate the practical impact of PCRI in real-world settings, we used PCRI to guide model selection for an internal MLLM-powered product during a closed beta launch (over 300 users, proprietary dataset). Models with near-zero PCRI were consistently preferred over negative-PCRI models, even when overall task accuracy was similar. User feedback highlighted greater reliability, consistency, and stability in outputs from models with higher context understanding, confirming that PCRI is predictive of real deployment success.

\paragraph{Conclusion.}
Our results confirm that negative PCRI is a %pathological, 
non-human-like model behavior, while human reasoning mostly benefits from additional context. PCRI thus provides a valuable, interpretable signal for practitioners seeking robust, trustworthy MLLMs for production use.

\subsubsection{PCRI on VQA Tasks}
\label{sec:appendix_vqa}
Visual Question Answering (VQA) tasks uniquely require integrating textual queries with visual details that may span broader contextual information across an image. Unlike MCQ or captioning tasks, where localized image patches consistently outperform full-image contexts (negative PCRI), our analysis reveals that VQA tasks exhibit a slight preference for full-image contexts, as indicated by uniformly small but positive PCRI scores (Figure~\ref{fig:vqa_avg}).

\paragraph{1. Mild Positive PCRI: Preference for Broader Context} Across  the models evaluated, PCRI values for VQA tasks remain modestly positive (typically between $0.003$ and $0.19$), signifying slight but consistent benefits from broader image contexts compared to localized patches. This pattern diverges notably from the strong negative PCRI observed in Captioning and MCQ tasks, underscoring VQA's inherent requirement for integrating more comprehensive visual information and relational context rather than isolated visual details.

\paragraph{2. Model-specific Variability} Significant variability exists across models. For instance, Pixtral-12B (PCRI=0.19 at $3\times3$) and Phi-3-Vision (PCRI=0.16 at $3\times3$) demonstrate the strongest preference for broader contexts.%, likely benefiting from integrated visual-textual reasoning strategies.
In contrast, InternVL2.5-2B and molmoE-1B-0924 show negligible PCRI scores (~0.01), indicating minimal differentiation between localized and full-context settings.%, possibly due to limited visual-textual alignment capabilities.

\paragraph{3. Impact of Patch Granularity} Increasing patch granularity from 
$2\times2$ to $3\times3$ generally results in moderately higher PCRI scores, indicating that models slightly prefer broader contexts (coarser granularity) over highly localized segments in VQA tasks. This suggests that global visual cues and contextual relationships become more salient and beneficial at lower granularities. However, the magnitude of these improvements remains moderate, implying that current MLLMs are already relatively effective at integrating visual information at coarser scales, and finer patches offer limited incremental advantages.

% \paragraph{3. Impact of Patch Granularity} Increasing patch granularity from $2\times2$ to $3\times3$ leads to slight increases in PCRI in most models, suggesting that, unlike Captioning and MCQ tasks, VQA tasks increasingly rely on global visual cues and contextual relationships that become more salient at lower patch granularities. However, the overall magnitude of these improvements remains moderate, suggesting that while additional granularity enhances context integration, current models already effectively leverage localized visual information at coarser scales.

% moderately finer segmentation generally degrades contextual understanding in VQA tasks. However, these increments remain modest, highlighting that beyond a certain granularity level, the benefits of further localized segmentation diminish, possibly due to the models' existing capacity to integrate relevant global visual contexts effectively. 

%  Increasing patch granularity from $2\times2$ to $3\times3$ generally results in moderately higher PCRI scores across most models, indicating an enhanced preference for broader contexts when finer patch segmentation is introduced. This suggests that, unlike Captioning and MCQ tasks, VQA models increasingly rely on global visual cues and contextual relationships that become more salient at higher patch granularities. However, the overall magnitude of these improvements remains moderate, suggesting that while additional granularity enhances context integration, current models already effectively leverage global visual information at coarser scales.

\paragraph{4. Contrasting VQA with MCQ and Captioning Tasks} Unlike Captioning and MCQ tasks, where models consistently prefer localized patches (negative PCRI scores), the VQA task demonstrates an inherent need for broader visual context integration. This difference likely arises from the open-ended nature of VQA tasks, requiring more comprehensive visual reasoning and understanding of inter-object relationships and semantic context. This finding aligns with recent insights in the literature emphasizing the importance of global visual context for effective VQA reasoning~\cite{jiang2022trips}%,antol2015vqa}.

\subsubsection{PCRI on MCQ Tasks}
\label{sec:appendix_mcq}

Multiple-Choice Question (MCQ) tasks require models to extract task-relevant information and choose a correct option to successfully solve the task. Figure~\ref{fig:mcq_avg} presents the PCRI across different models for $2 \times 2$ and $3 \times 3$ patch-based inputs, averaged over multiple MCQ datasets.%, while Table \ref{tab:mcq_detail} provides the details across each dataset

Across all models, PCRI remains negative, indicating that patch-based inputs consistently outperform full-image contexts. This suggests that global image representations potentially introduce unnecessary distractions or miss capturing necessary information during image encoding , diluting attention mechanisms and reducing task-specific accuracy. 

We also observe that models generally perform better at $n=3$ (more localized patches) compared to $n=2$, reinforcing the hypothesis that MLLMs struggle to process global context effectively. Certain architectures, such as Molmo-1B, NVLM, and Janus-Pro, show more significant improvements in localized patch-based settings, implying that these models are particularly vulnerable to irrelevant background information. Conversely, larger-scale models like InternVL2.5-26B and Qwen2-VL-72B exhibit more stable PCRI, suggesting that increased capacity may improve context handling, though not entirely eliminate sensitivity to global context.

\subsubsection{PCRI on Yes/No Question Answering Tasks}
\label{sec:appendix_yn}

Yes/No question-answering tasks present a distinct challenge for multimodal models, as they often require binary reasoning over image content. Unlike open-ended VQA tasks, Y/N datasets tend to emphasize disambiguation of objects, attributes, or relationships within an image, making them an important benchmark for evaluating the impact of full-image versus localized context processing.

\paragraph{1. Localized Patches Improve Y/N Answering}  
Across nearly all models, we observe consistently negative PCRI values, indicating that the models perform better when using localized patches rather than full-image input. This suggests that full-image representations introduce unnecessary context, leading to increased ambiguity in binary classification tasks.

\paragraph{2. Patch Size Influences Performance Gains}  
When comparing \(2 \times 2\) and \(3 \times 3\) patches, the latter consistently yields lower PCRI values, meaning greater patch granularity improves performance. This supports the hypothesis that more focused image regions help models resolve Y/N questions by minimizing distractions.

\paragraph{3. Relationship Between Model Scale and Context Handling}  
Larger models, such as Pixtral-12B, demonstrate a more stable PCRI, suggesting that scaling helps manage full-image context better. However, even for high-capacity models, localized patches still provide a performance boost, indicating that current attention mechanisms remain suboptimal for global reasoning in binary tasks.

These results further emphasize the need for adaptive attention filtering, where models dynamically adjust the level of contextual information they consider based on task requirements.

\subsubsection{Model Level Analysis}

\label{sec:appendix_model_level_analysis}
\paragraph{Phi-3 Models:}
Phi-3 models consistently display moderate negative PCRI scores across tasks. Despite their compact size, Phi-3 architectures leverage efficient attention and optimized training strategies, mitigating severe performance degradation at higher granularities. However, their robustness still trails behind larger architectures like InternVL-26B and Qwen2-VL-72B, reflecting inherent capacity limitations for managing extensive visual context~\cite{abdin2024phi}.

\paragraph{Janus-Pro Models:}
Janus-Pro shows significant negative PCRI values across all tasks, notably severe in MCQ and captioning. This indicates substantial sensitivity to global context, attributable to its dual visual encoder approach, separately optimized for understanding and generation. Although Janus-Pro excels at specific multimodal generation tasks, its fragmented encoding strategy negatively impacts robustness in holistic scene comprehension, highlighting trade-offs in encoder specialization~\cite{chen2025janus}.

\paragraph{Ovis Models:}
Ovis variants present mixed PCRI scores, with smaller versions (Llama3.2-3B) heavily context-sensitive, while larger ones (Gemma2-9B) demonstrate improved robustness. The structured visual embedding strategy employed by Ovis provides a clear theoretical advantage for cross-modal embedding alignment, yet smaller variants still struggle with overwhelming contextual information due to limited embedding capacity. Larger Ovis models better leverage structured embeddings, balancing detailed visual perception with enhanced robustness~\cite{lu2024ovis}.

\paragraph{Pixtral Models:}
Pixtral-12B exhibits notably high PCRI sensitivity, particularly in Yes/No and captioning tasks, suggesting challenges in effectively processing global contexts despite advanced ROPE-2D encoding strategies. This sensitivity highlights inherent trade-offs associated with high-resolution, multi-image reasoning, where detailed attention enhances local perception at the cost of global context integration~\cite{agrawal2024pixtral12b}.
\begin{figure*}[!th]
    \centering
    \includegraphics[width=\linewidth]{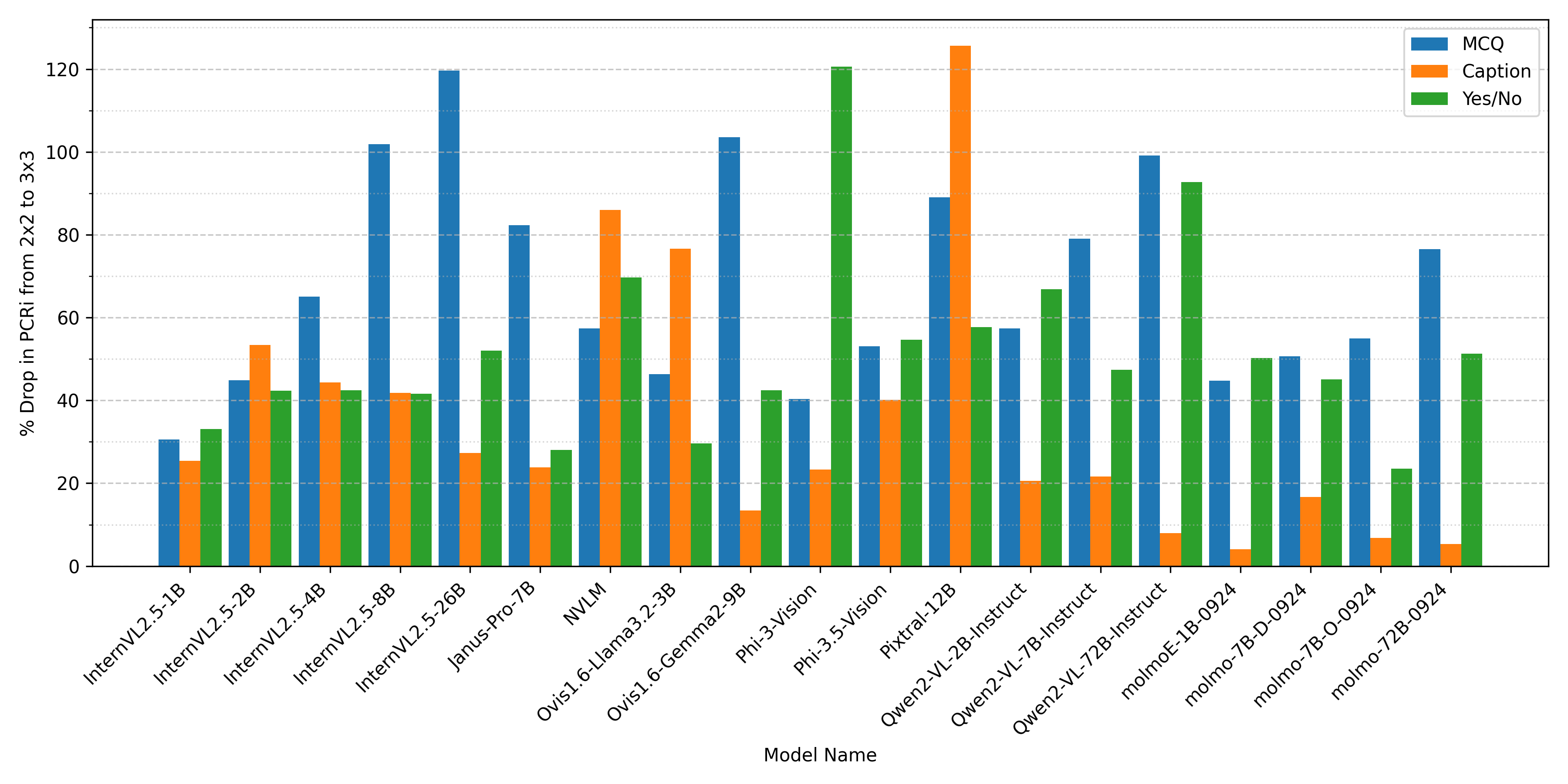}
    \caption{Percentage change in PCRI from $2 \times 2$ to $3 \times 3$ patches across MCQ, Caption, and Y/N datasets. A higher drop in PCRI suggests greater sensitivity to patch granularity, while a smaller drop indicates better stability in handling full-image contexts.}
    % \vspace{-2em}
    \label{fig:drop_perf}
\end{figure*}

\subsubsection{Impact of Patch Granularity on PCRI Across Tasks}
\label{sec:appendix_granularity}
To further explore how varying visual granularities influence context robustness, we analyze the relative changes in PCRI scores when moving from $2 \times 2$ to finer-grained $3 \times 3$ patches across three primary task types: Multiple-choice QA (MCQ), Captioning, and Yes/No (Y/N) classification (Figure~\ref{fig:drop_perf}). This analysis complements the absolute PCRI evaluations (Figures \ref{fig:mcq_avg}, \ref{fig:caption_avg}, \ref{fig:yn_avg}) from our main results, highlighting important subtleties in the interaction between visual granularity and model performance. Performance across different models and datasets can be found in Table~\ref{tab:mcq_detail}, \ref{tab:yn_detail}, \ref{tab:vqa_detail}, and \ref{tab:caption_detail}.

\paragraph{1. Interpreting Relative Drop in PCRI} Figure~\ref{fig:drop_perf} depicts the percentage drop in PCRI, measuring the magnitude of performance change when visual context granularity increases. Importantly, these relative changes must be interpreted in conjunction with absolute PCRI values to avoid misleading conclusions. A higher percentage drop may not necessarily indicate poor absolute robustness if initial PCRI magnitudes are small.

\paragraph{2. MCQ Tasks: High Relative Drops with Moderate Absolute Impact}
MCQ tasks exhibit the largest average relative PCRI drop (68.24\%), suggesting that further granularity greatly enhances the advantage of localized patches over full images. For instance, InternVL2.5-26B shows an extreme relative drop (119.6\%), yet its absolute PCRI remains relatively moderate ($-0.036$ at $2\times2$ to $-0.079$ at $3\times3$). Similarly, Qwen2-VL-72B-Instruct displays a 99.1\% relative drop but maintains low absolute PCRI ($-0.013$ to $-0.027$). This highlights that hierarchical models exhibit substantial robustness to image and context granularity, with a slight improvement in performance on MCQ tasks as granularity increases.

\paragraph{3. Captioning Tasks: Moderate Relative Drops with Significant Absolute PCRI}
Captioning tasks have a smaller average relative PCRI drop (34.95\%) compared to MCQ, but their absolute PCRI magnitudes are notably higher. For example, InternVL2.5-2B experiences a moderate relative change (53.4\%) but displays significant absolute PCRI values ($-0.492$ at $2\times2$ to $-0.755$ at $3\times3$). Thus, captioning tasks consistently emphasize localized context, as previously noted, and even modest increases in granularity substantially amplify models' preference for localized patches, due to inherent task characteristics.

\paragraph{4. Yes/No Tasks: Balanced Relative and Absolute PCRI Values}
Yes/No classification tasks demonstrate intermediate behavior, with an average relative PCRI drop of 52.16\% and moderate absolute PCRI magnitudes (typically below $-0.15$ at $3\times3$). Notably, some models like Phi-3-Vision (120.6\% relative drop) and Qwen2-VL-72B (92.7\% relative drop) exhibit significant sensitivity to granularity, indicating that while binary classification generally relies on simpler visual cues, specific architectural choices significantly influence robustness.

\paragraph{5. Architectural Implications and Context Sensitivity} The variability observed underscores the complexity of interpreting PCRI in context. Models with sophisticated hierarchical attention (e.g., InternVL and Qwen2-VL larger models) tend to achieve strong absolute robustness, despite higher relative sensitivity to granularity. Conversely, smaller or simpler architectures experience pronounced absolute and relative degradation, highlighting critical vulnerabilities in their visual encoding and attention mechanisms.

% \paragraph{6. Practical Recommendations for Model Deployment} Practitioners should carefully balance relative and absolute PCRI interpretations. High absolute PCRI values indicate substantial overall context sensitivity, directly affecting real-world reliability. High relative drops, while informative, primarily reflect the additional benefit models receive from further localized context. Optimal model deployment decisions should thus prioritize absolute robustness first, then leverage relative insights to fine-tune visual encoding strategies based on task-specific requirements.

\subsection{Illustrative Examples of Context Granularity}
\label{sec:example_PCRI}

To qualitatively illustrate how PCRI captures model sensitivity to context, Table~\ref{tab:realworldqa_pcri_examples} presents examples from RealWorldQA. For each question, we show the original image, its $2 \times 2$ and $3 \times 3$ patch splits, and highlight each patch: \textbf{green boundaries} indicate that the model answered correctly using only that patch, while \textbf{red boundaries} indicate incorrect predictions at that context level.

These examples highlight several key points:

\begin{table*}[!ht]
\centering
\renewcommand{\arraystretch}{1.2} % Adjust row height
\setlength\dashlinedash{0.5pt}    % Dash length
\setlength\dashlinegap{1pt}       % Gap between dashes
\setlength\arrayrulewidth{0.8pt}  % Thickness of outer border

\vspace{0.5em}

\scalebox{0.85}{
\begin{tabular}{|m{3cm}  % Question - justified
                |>{\centering\arraybackslash}m{3.5cm}  % Original image
                |>{\centering\arraybackslash}m{3.5cm}  % 2x2 Split
                |>{\centering\arraybackslash}m{3.5cm}  % 3x3 Split
                |m{3cm}                                % Answerable with - justified
                |>{\centering\arraybackslash}m{2cm}|}  % Right-most column (if needed)
\hline
\textbf{Question} & \textbf{Original} & \textbf{2×2 Split} & \textbf{3×3 Split} & \textbf{Answerable with} \\
\hline

Is the crosswalk sign active? Please answer directly with a single word or number. &
\includegraphics[height=2cm]{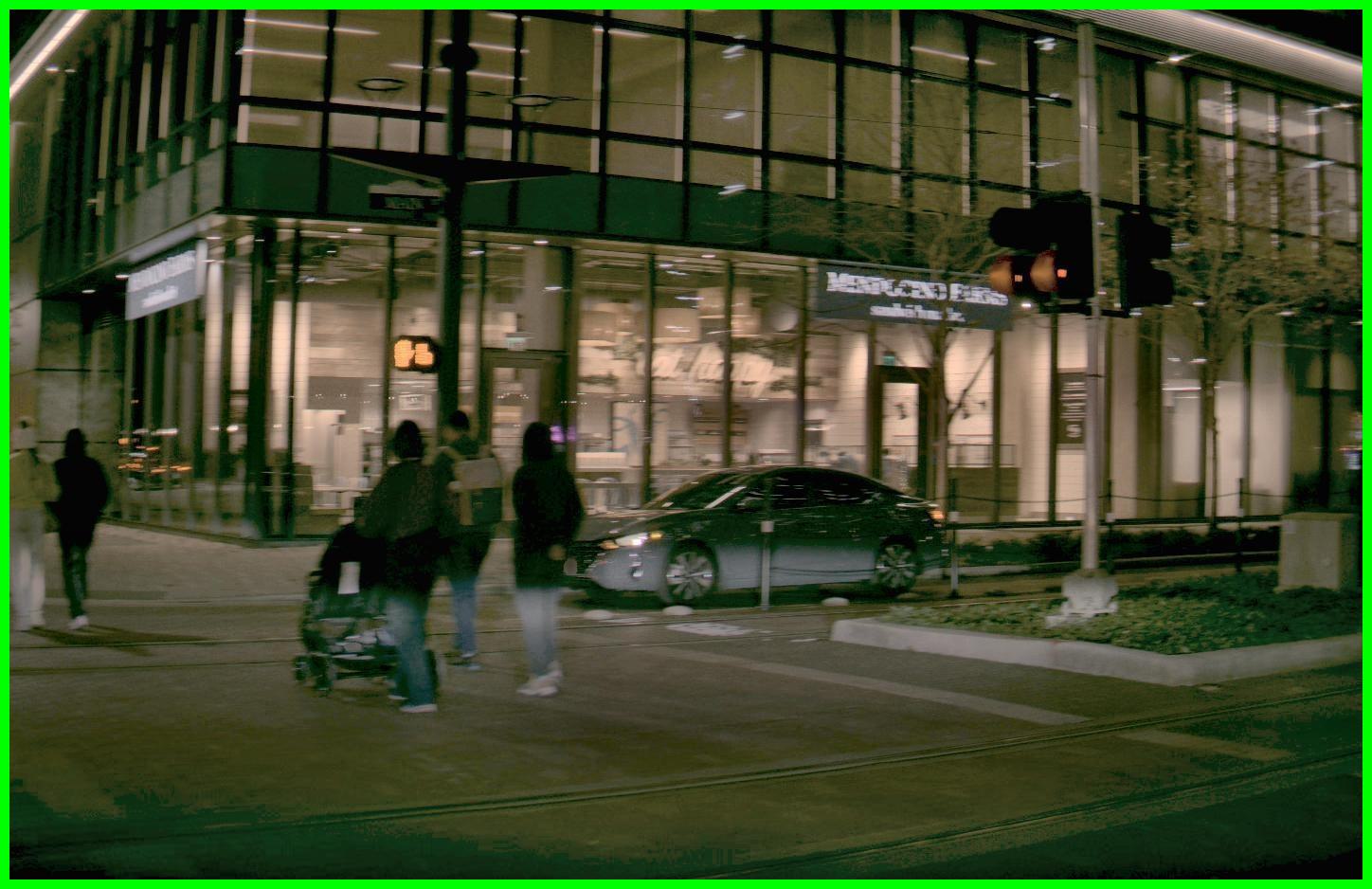} &
\includegraphics[height=2cm]{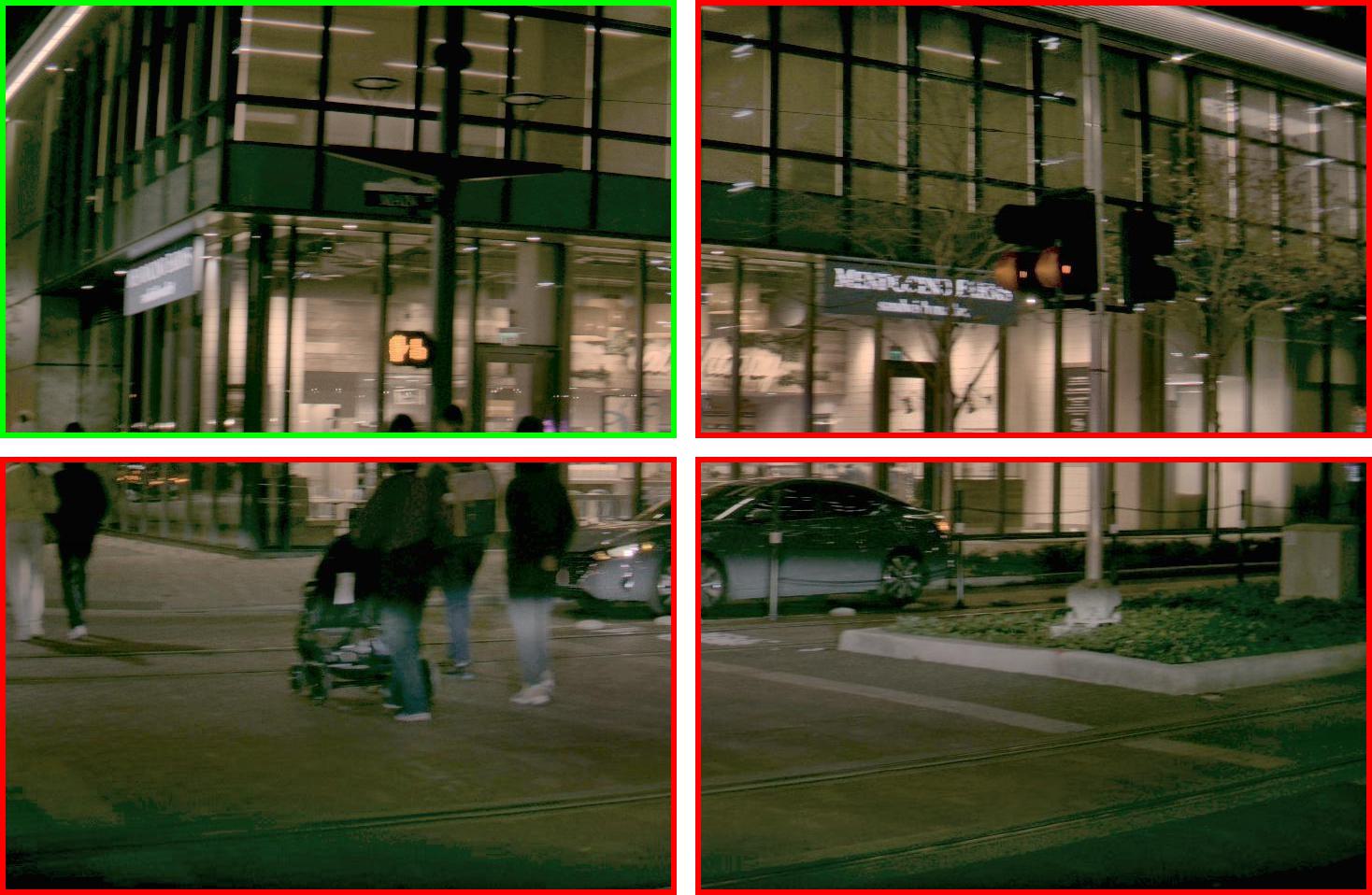} &
\includegraphics[height=2cm]{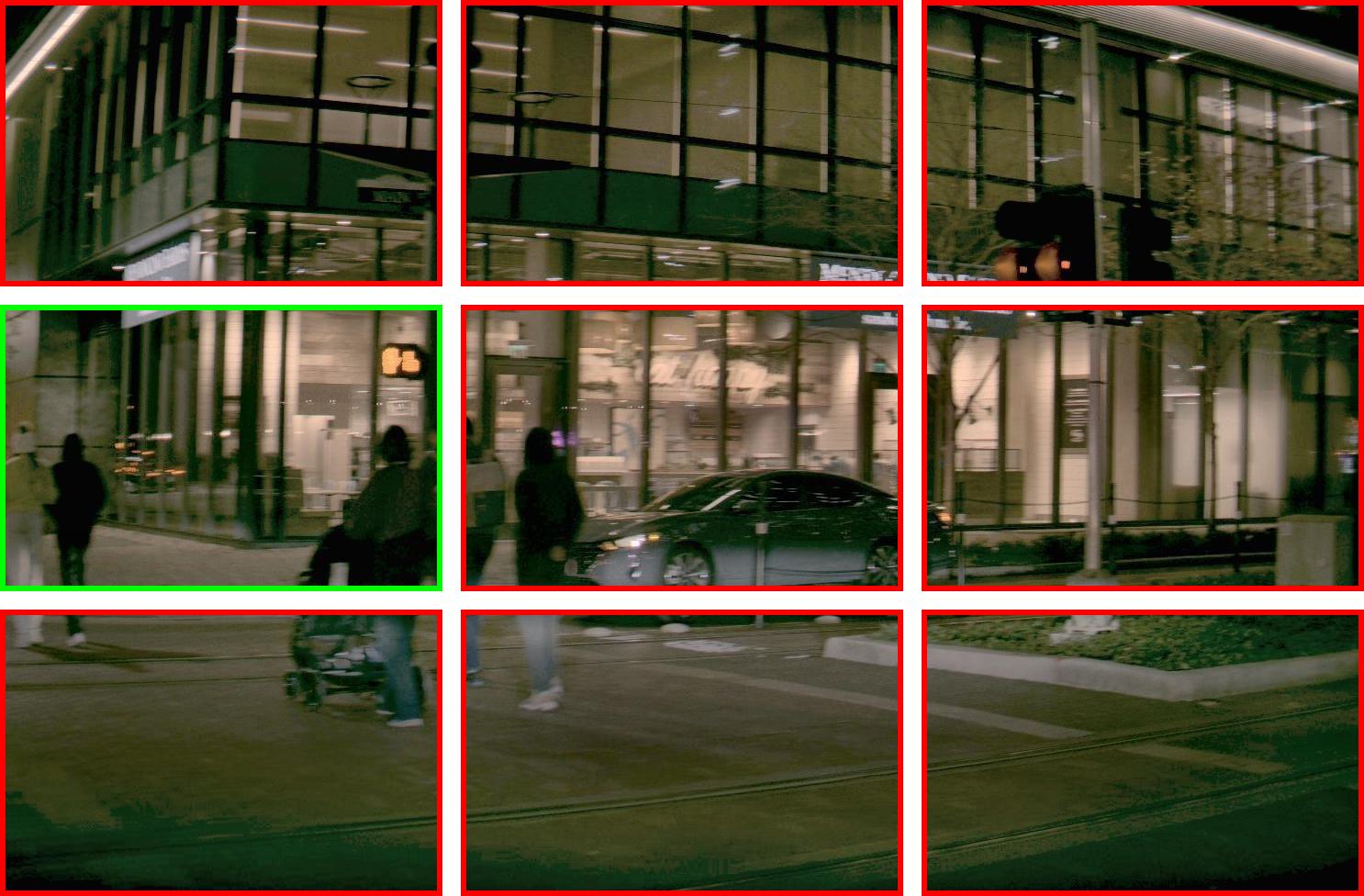} &
Full image, 2×2 split (row 1, column 1), 3×3 split (row 2, column 1) \\
\hline

Is there a stop sign in this image? Please answer directly with a single word or number. &
\includegraphics[height=2cm]{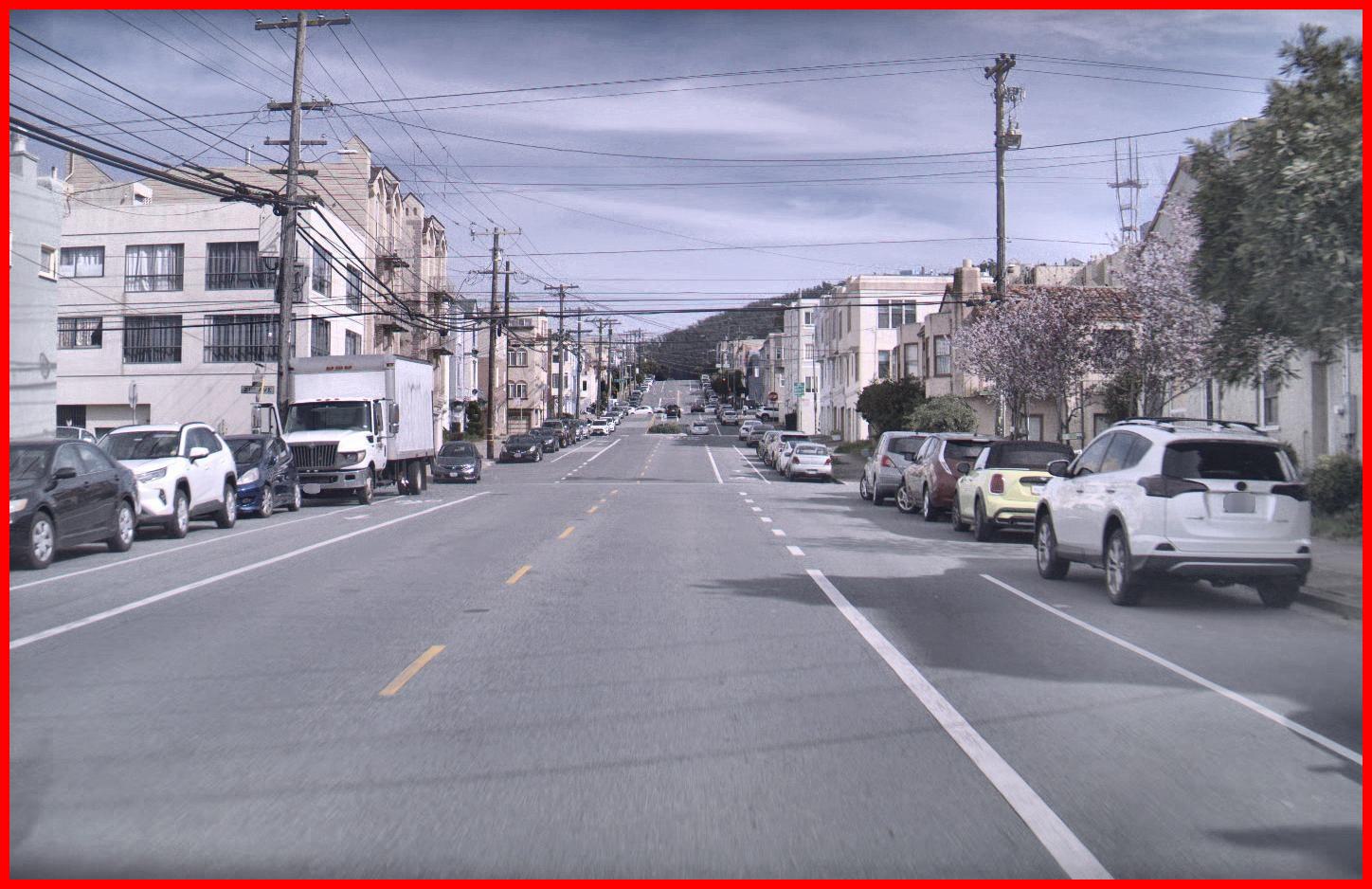} &
\includegraphics[height=2cm]{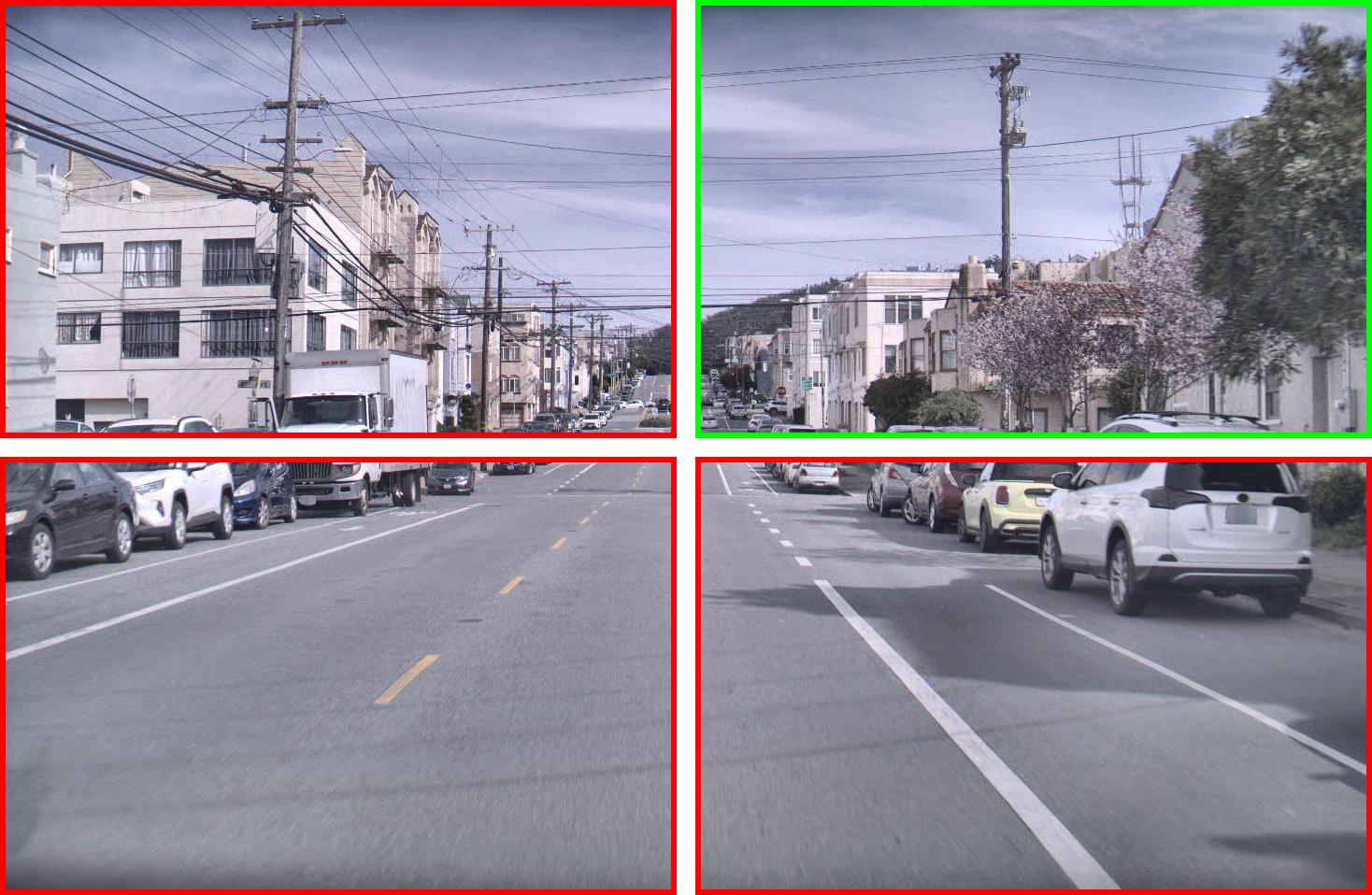} &
\includegraphics[height=2cm]{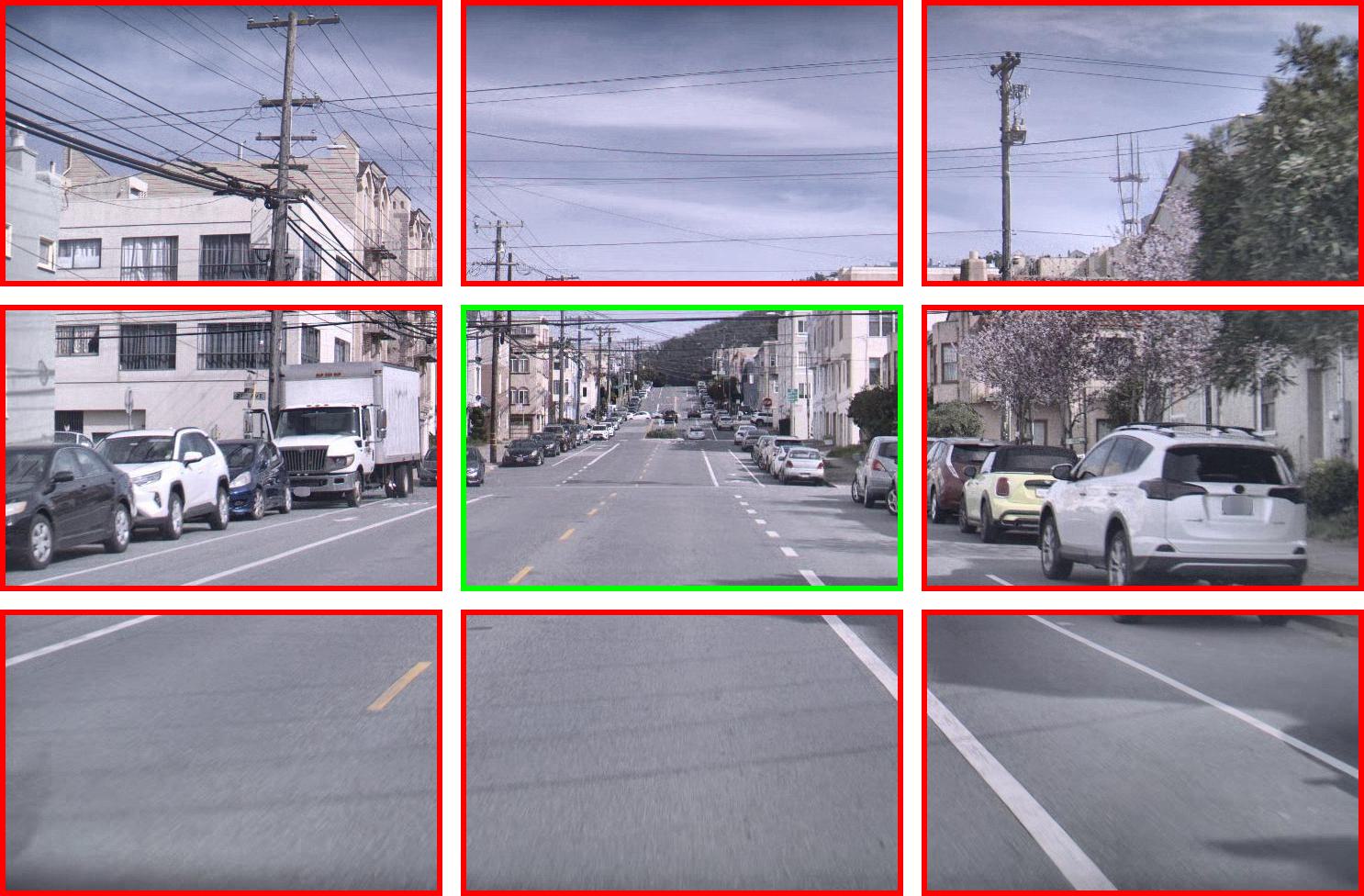} &
2×2 split (row 1, column 2), 3×3 split (row 2, column 2) \\
\hline

How many pedestrians are there? Please answer directly with a single word or number. &
\includegraphics[height=2cm]{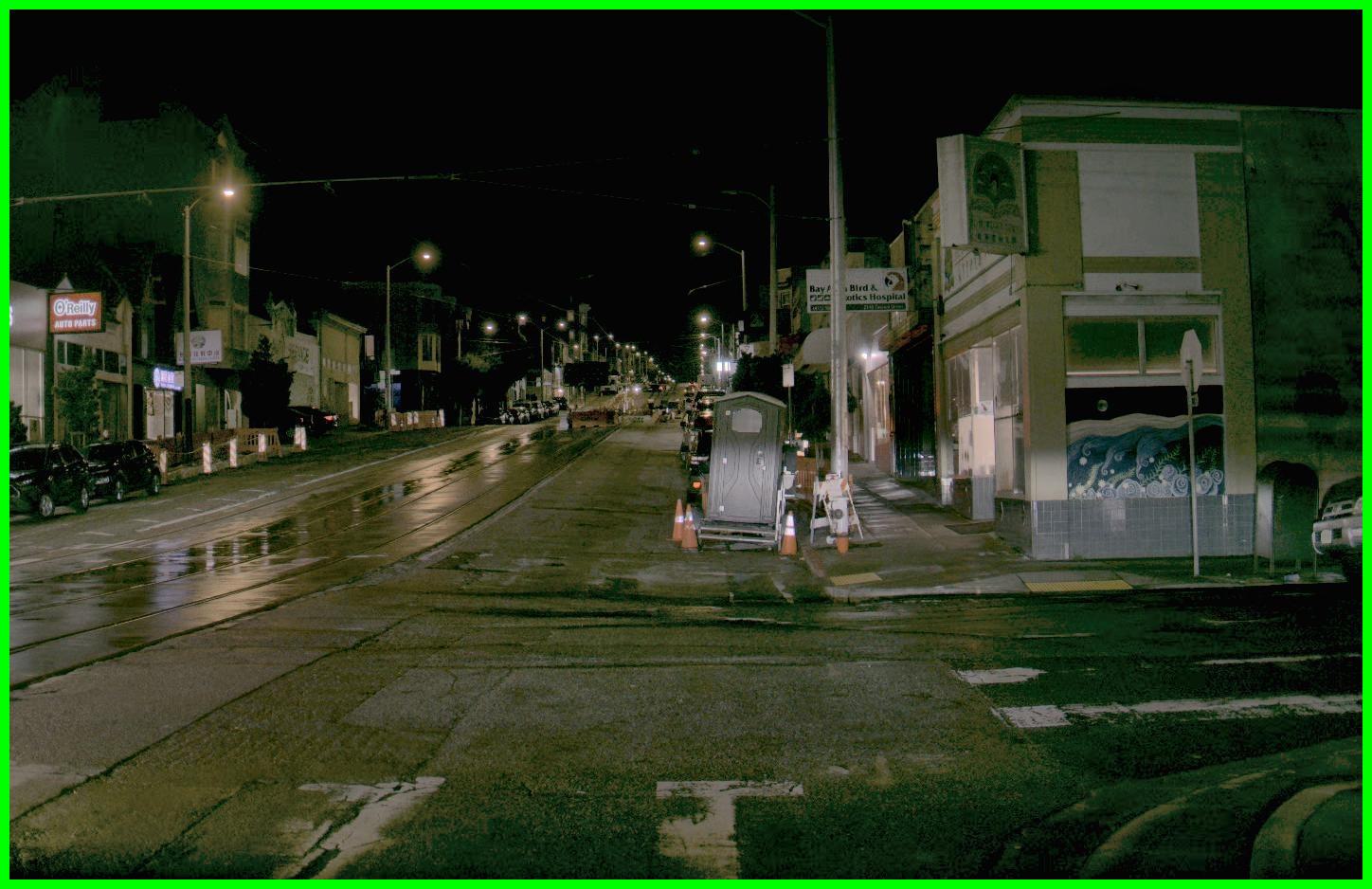} &
\includegraphics[height=2cm]{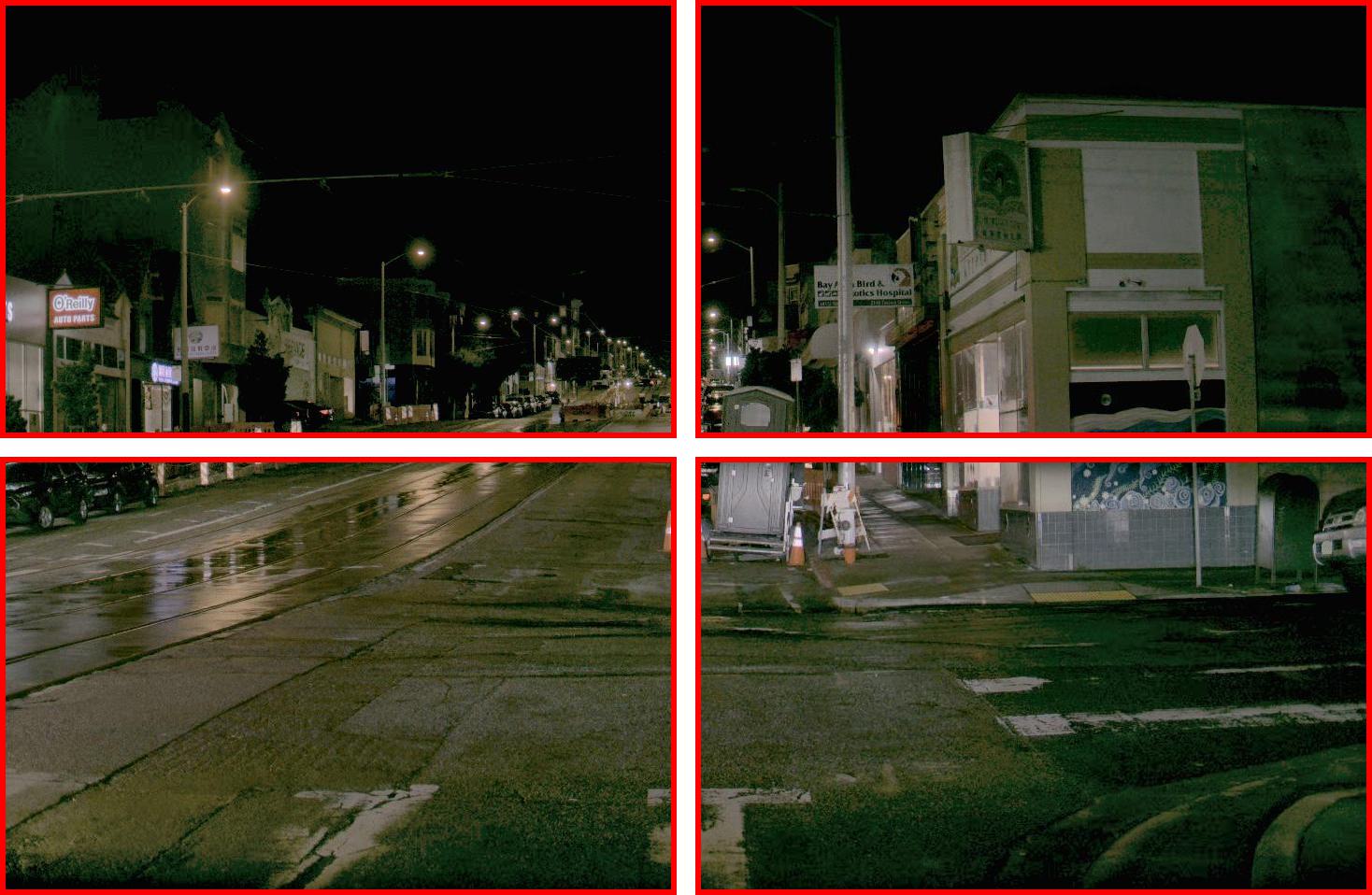} &
\includegraphics[height=2cm]{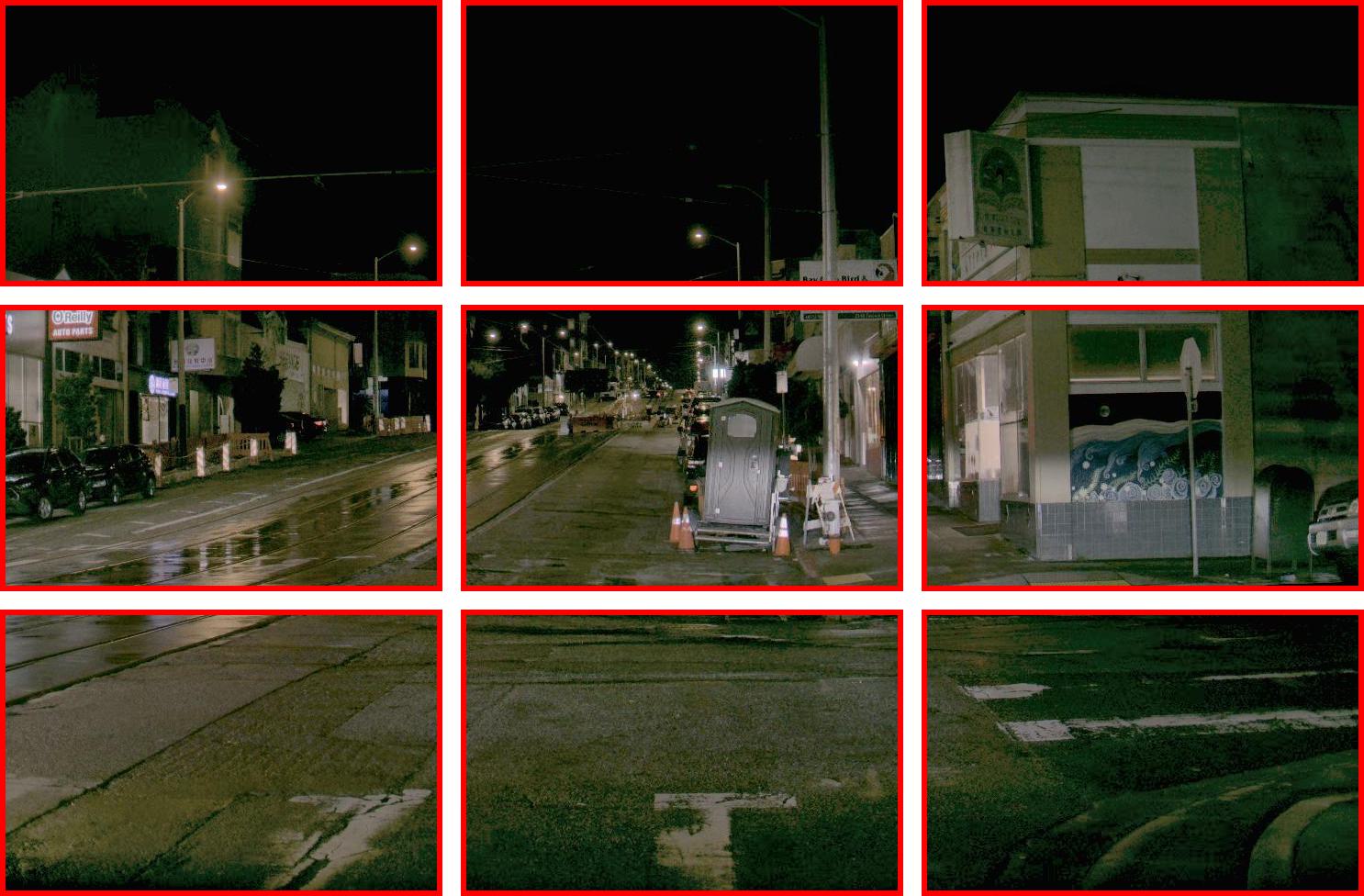} &
Full image \\
\hline

\end{tabular}
}
\caption{
Illustrative examples from the RealWorldQA dataset demonstrating how context granularity (full image vs. 2×2 and 3×3 splits) affects answerability and model accuracy. \textcolor{green}{Green boundaries} indicate correct answers at that patch/context; \textcolor{red}{Red boundaries} indicate incorrect predictions.
}
\label{tab:realworldqa_pcri_examples}
\vspace{-1em}
\end{table*}

\begin{itemize}
    \item \textbf{Varied context requirements:} Some questions can be answered correctly from a single local patch (e.g., detection of a stop sign), while others require more global context or integration across regions (e.g., counting pedestrians).
    \item \textbf{PCRI as a diagnostic tool:} By analyzing which patches yield correct versus incorrect predictions, practitioners can diagnose local and global reasoning capabilities, as well as identify where context brittleness emerges.
    \item \textbf{Practical implications:} Such analysis informs targeted dataset curation, model development, and real-world deployment, by matching model selection and training to the true context requirements of end-user tasks.
\end{itemize}

Overall, these qualitative examples reinforce the value of PCRI as a fine-grained, interpretable measure of context robustness in vision-language models.

% \begin{table}[h]
%     \centering
%     \includegraphics[width=\linewidth]{latex/images/pcri_realworldqa_examples.png}
%     \caption{Illustrative examples from RealWorldQA showing how context granularity (full image, $2 \times 2$, $3 \times 3$ splits) affects model accuracy. \textbf{Green boundaries} indicate correct answers at that patch/context; \textbf{red boundaries} indicate incorrect predictions.}
%     \label{tab:pcri_realworldqa_examples}
% \end{table}

\section{Practical Implications and Usage Guide}
\label{sec:practitioner_appendix}

% \subsection{PCRI in Applied Workflows}
The Patch Context Robustness Index (PCRI) offers a concrete, actionable tool for practitioners designing, auditing, and maintaining multimodal models for real-world applications. PCRI can be integrated throughout the MLLM lifecycle to:
\begin{itemize}
    \item \textbf{Benchmark Model Robustness}: Rank candidate models on context sensitivity before deployment, ensuring chosen models remain reliable in environments with background clutter, occlusions, or incomplete views.
    \item \textbf{Monitor Deployed Systems}: Track PCRI over time to detect emerging vulnerabilities as data distributions shift, e.g., in dynamic environments or after retraining.
    \item \textbf{Auditing and Debugging}: Use PCRI to identify tasks or datasets where a model is brittle (negative or high PCRI), guiding further data collection or model refinement.
    \item \textbf{Compare Training Strategies}: Evaluate the effect of architectural choices or pretraining schemes on context robustness, providing a rigorous basis for model design.
\end{itemize}

\subsection{Interpretation Guide}

PCRI values provide actionable signals:%(redundant)
\begin{itemize}
    \item \textbf{PCRI $\approx 0$}: Model is robust; context granularity does not affect performance. Suitable for deployment in unpredictable visual conditions.
    \item \textbf{PCRI $> 0$}: Model requires global context (e.g., scene-level tasks); patch-only input insufficient.
    \item \textbf{PCRI $< 0$}: Model exploits local cues but fails globally, indicative of shortcut or brittle behavior; not human-like and poses deployment risk.
    \item \textbf{Undefined/large values}: Denominator is too small or zero; metric unreliable for that task/model combination.
\end{itemize}

\noindent\textit{Best practices:} Use PCRI with other robustness metrics. Investigate negative PCRI to diagnose spurious correlations. Monitor PCRI regularly in production as model and data evolve.

\subsection{Use-Cases}

PCRI can be integrated at multiple stages of AI product development, deployment, and maintenance, offering value for robustness, reliability, and transparency in diverse settings:

\begin{itemize}
    \item \textbf{Retail Product Search and E-Commerce:} Cluttered, dynamic shelves and varied camera angles can introduce substantial background noise and distractors. PCRI enables teams to benchmark and select models that sustain high retrieval or classification accuracy despite irrelevant objects, reducing false positives and improving user trust in product recommendations, search and survellaince.

    \item \textbf{Assistive Technology for Accessibility:} For screen readers, object recognizers, and navigation aids used by visually impaired individuals, input images are often cropped, occluded, over-zoomed or partially visible. PCRI ensures selected models are robust in such scenarios, decreasing risk of missed cues or misleading outputs, and supporting safer user experiences in everyday environments.

    \item \textbf{Autonomous Vehicles and Robotics:} Changing backgrounds (construction, seasonal foliage, weather conditions, or dynamic obstacles) can degrade MLLM performance. By tracking PCRI over time, operators can identify when models become brittle to new environmental context, triggering targeted data augmentation, model retraining, or human-in-the-loop overrides before safety-critical failures.

    \item \textbf{Industrial Inspection and Quality Control:} Automated inspection systems (e.g., for manufacturing defects) must distinguish true faults from distracting background patterns or partial occlusions. PCRI supports the benchmarking of new models for robustness against such nuisance variation, guiding dataset augmentation and QA pipelines.

    \item \textbf{Content Moderation and Safety:} Social media and online platforms face adversarial attempts to evade detection via occlusion, cropping, or clutter. PCRI can flag models that are sensitive to such manipulations, helping design systems that maintain detection accuracy in the presence of adversarial context modification.

    \item \textbf{Medical Imaging and Diagnostics:} Context brittleness in radiology or pathology images can lead to missed findings or false alarms due to artifacts, cropping, or scanner noise. PCRI helps validate models on edge cases where only local detail is diagnostic, supporting higher reliability for clinical deployment and regulatory clearance.

    \item \textbf{Continuous Model Monitoring and Drift Detection:} In production, real-world data distributions evolve. Integrating PCRI into monitoring dashboards enables early detection of performance drift due to novel backgrounds or scene elements, supporting proactive retraining and minimizing negative user impact.

    \item \textbf{Model Regression Testing and Compliance Audits:} PCRI offers a standardized, quantitative metric for comparing successive model versions on context robustness, providing a clear “go/no-go” signal for deployment. Including PCRI in model cards or audit logs supports regulatory compliance and transparent documentation for stakeholders.

    \item \textbf{Benchmark and Dataset Curation:} PCRI can highlight underrepresented context challenges in existing benchmarks. Dataset designers can use PCRI analysis to guide new data collection—adding samples with cluttered, ambiguous, or challenging backgrounds to improve model generalization.

    \item \textbf{Internal Model Selection and Beta Testing:} As shown in our production beta launch (see Section~\ref{sec:human_study_appendix}), models with higher (near-zero) PCRI delivered improved user feedback and reliability, even when overall accuracy was matched, highlighting PCRI’s value for practitioner-facing decision-making.
\end{itemize}

\subsection{Limitations and Caveats}
\begin{itemize}
    \item \textbf{Small denominators:} PCRI is unstable or undefined if few images are correct globally. Exclude such tasks or use bootstrapped intervals.
    \item \textbf{Patch granularity:} Excessively large $n$ results in tiny, meaningless patches; use $n=2$ or $n=3$.
    \item \textbf{Metric agnosticism:} PCRI can use any base metric (accuracy, F1, etc.) as long as patch/whole scores are defined.
    \item \textbf{Independence:} PCRI ignores spatial dependencies between patches; future work may address this.
\end{itemize}

\begin{table*}[ht]
    \centering
    \label{tab:model_comparison}
    \renewcommand{\arraystretch}{1.2} % Adjust row height
    \setlength\dashlinedash{0.85pt}    % Dash length
    \setlength\dashlinegap{1pt}       % Gap between dashes
    \setlength\arrayrulewidth{0.8pt}  % Thickness of outer border
    
    % Adjust column width manually
    \scalebox{0.5}{ 
    \begin{tabular}{|l|p{1.8cm}:p{1.8cm}|p{1.8cm}:p{1.8cm}|p{1.8cm}:p{1.8cm}|p{1.8cm}:p{1.8cm}|p{1.8cm}:p{1.8cm}|p{1.8cm}:p{1.8cm}|}
        \hline
        \textbf{Model} & \multicolumn{2}{c|}{\textbf{AI2D}} & \multicolumn{2}{c|}{\textbf{BLINK}} & \multicolumn{2}{c|}{\textbf{MMMU}} & \multicolumn{2}{c|}{\textbf{MMStar}} & \multicolumn{2}{c|}{\textbf{RealWorldQA}} & \multicolumn{2}{c|}{\textbf{ScienceQA}} \\
        \cline{2-13}
        & PCRI$_{n=2}$ & PCRI$_{n=3}$ & PCRI$_{n=2}$ & PCRI$_{n=3}$ & PCRI$_{n=2}$ & PCRI$_{n=3}$ & PCRI$_{n=2}$ & PCRI$_{n=3}$ & PCRI$_{n=2}$ & PCRI$_{n=3}$ & PCRI$_{n=2}$ & PCRI$_{n=3}$ \\
        \hline
        InternVL2\_5-1B & -0.17 & -0.23 & -0.05 & -0.23 & -0.21 & -0.04 & -0.22 & -0.29 & -0.05 & -0.29 & -0.31 & -0.03 \\
        InternVL2\_5-26B & -0.05 & -0.03 & -0.03 & -0.09 & -0.07 & -0.06 & -0.07 & -0.16 & -0.03 & -0.13 & -0.13 & -0.05 \\
        InternVL2\_5-2B & -0.09 & -0.38 & -0.01 & -0.11 & -0.20 & -0.04 & -0.13 & -0.45 & 0.01 & -0.23 & -0.28 & -0.02 \\
        InternVL2\_5-4B & -0.08 & -0.15 & -0.04 & -0.11 & -0.17 & -0.07 & -0.10 & -0.25 & -0.04 & -0.20 & -0.25 & -0.06 \\
        InternVL2\_5-8B & -0.05 & -0.01 & -0.03 & -0.09 & -0.16 & -0.07 & -0.08 & -0.10 & -0.03 & -0.17 & -0.18 & 0.05 \\
        Janus-Pro-7B & -0.12 & -0.50 & -0.02 & -0.19 & -0.24 & -0.01 & -0.19 & -0.70 & 0.02 & -0.34 & -0.62 & -0.07 \\
        NVLM & -0.16 & -0.47 & -0.08 & -0.13 & -0.17 & -0.05 & -0.22 & -0.70 & -0.15 & -0.24 & -0.29 & -0.10 \\
        Ovis1.6-Gemma2-9B & -0.04 & -0.20 & -0.02 & -0.04 & -0.12 & -0.04 & -0.08 & -0.33 & -0.08 & -0.15 & -0.18 & -0.01 \\
        Ovis1.6-Llama3.2-3B & -0.13 & -0.45 & -0.10 & -0.23 & -0.17 & -0.04 & -0.17 & -0.70 & -0.14 & -0.32 & -0.24 & -0.07 \\
        Phi-3-Vision & -0.09 & -0.24 & -0.08 & -0.24 & -0.26 & -0.04 & -0.12 & -0.35 & -0.11 & -0.33 & -0.33 & 0.02 \\
        Phi-3.5-Vision & -0.10 & -0.04 & -0.06 & -0.27 & -0.31 & 0.02 & -0.13 & -0.10 & -0.09 & -0.37 & -0.45 & -0.01 \\
        Pixtral-12B & -0.18 & -0.12 & -0.12 & -0.17 & -0.14 & -0.13 & -0.27 & -0.37 & -0.28 & -0.38 & -0.23 & -0.17 \\
        Qwen2-VL-2B-Instruct & -0.11 & -0.20 & -0.04 & -0.26 & -0.17 & -0.07 & -0.16 & -0.37 & -0.04 & -0.39 & -0.27 & -0.12 \\
        molmo-72B-0924 & -0.12 & -0.31 & -0.02 & -0.03 & -0.09 & -0.02 & -0.16 & -0.43 & -0.07 & -0.11 & -0.17 & -0.12 \\
        molmo-7B-D-0924 & -0.12 & -0.33 & -0.07 & -0.13 & -0.10 & 0.00 & -0.15 & -0.41 & -0.12 & -0.23 & -0.20 & -0.03 \\
        molmo-7B-O-0924 & -0.13 & -0.27 & -0.08 & -0.18 & -0.11 & -0.03 & -0.17 & -0.40 & -0.12 & -0.29 & -0.20 & -0.04 \\
        molmoE-1B-0924 & -0.22 & -0.38 & -0.17 & -0.30 & -0.22 & -0.04 & -0.32 & -0.52 & -0.27 & -0.46 & -0.31 & -0.08 \\
        \hline
    \end{tabular}
    }
    \caption{PCRI MCQ scores for different models across datasets.}
    \vspace{-1em}
    \label{tab:mcq_detail}
\end{table*}

\begin{table*}[ht]
    \centering
    \label{tab:model_comparison_2}
    \renewcommand{\arraystretch}{1.2} % Adjust row height
    \setlength\dashlinedash{0.85pt}    % Dash length
    \setlength\dashlinegap{1pt}       % Gap between dashes
    \setlength\arrayrulewidth{0.8pt}  % Thickness of outer border
    
    % Adjust column width manually
    \scalebox{0.7}{ 
    \begin{tabular}{|l|p{1.8cm}:p{1.8cm}|p{1.8cm}:p{1.8cm}|p{1.8cm}:p{1.8cm}|p{1.8cm}:p{1.8cm}|}
        \hline
        \textbf{Model} & \multicolumn{2}{c|}{\textbf{AMBER}} & \multicolumn{2}{c|}{\textbf{HallusionBench}} & \multicolumn{2}{c|}{\textbf{MME}} & \multicolumn{2}{c|}{\textbf{POPE}} \\
        \cline{2-9}
        & PCRI$_{n=2}$ & PCRI$_{n=3}$ & PCRI$_{n=2}$ & PCRI$_{n=3}$ & PCRI$_{n=2}$ & PCRI$_{n=3}$ & PCRI$_{n=2}$ & PCRI$_{n=3}$ \\
        \hline
        InternVL2\_5-1B & -0.01 & -0.27 & -0.06 & -0.05 & -0.03 & -0.35 & -0.10 & -0.06 \\
        InternVL2\_5-26B & -0.01 & -0.10 & -0.02 & -0.06 & -0.03 & -0.15 & -0.04 & -0.07 \\
        InternVL2\_5-2B & -0.06 & -0.22 & -0.05 & -0.05 & 0.04 & -0.27 & -0.08 & -0.06 \\
        InternVL2\_5-4B & -0.01 & -0.17 & -0.05 & -0.05 & -0.03 & -0.26 & -0.06 & -0.07 \\
        InternVL2\_5-8B & -0.02 & -0.16 & -0.03 & -0.05 & -0.04 & -0.22 & -0.05 & -0.05 \\
        Janus-Pro-7B & -0.01 & -0.22 & -0.05 & -0.05 & 0.00 & -0.28 & -0.07 & -0.06 \\
        NVLM & -0.02 & -0.27 & -0.07 & -0.04 & -0.04 & -0.49 & -0.09 & -0.05 \\
        Ovis1.6-Gemma2-9B & -0.03 & -0.19 & -0.08 & -0.05 & -0.05 & -0.26 & -0.12 & -0.06 \\
        Ovis1.6-Llama3.2-3B & -0.04 & -0.29 & -0.10 & -0.07 & -0.06 & -0.37 & -0.13 & -0.08 \\
        Phi-3-Vision & -0.03 & -0.09 & -0.09 & -0.05 & -0.01 & -0.16 & -0.20 & -0.06 \\
        Phi-3.5-Vision & -0.04 & -0.14 & -0.11 & -0.08 & 0.02 & -0.22 & -0.14 & -0.09 \\
        Pixtral-12B & -0.17 & -0.36 & -0.06 & -0.18 & -0.21 & -0.65 & -0.09 & -0.26 \\
        Qwen2-VL-2B-Instruct & -0.07 & -0.22 & -0.11 & -0.06 & -0.02 & -0.35 & -0.12 & -0.07 \\
        Qwen2-VL-72B-Instruct & -0.01 & -0.10 & -0.03 & -0.05 & -0.04 & -0.22 & -0.05 & -0.07 \\
        Qwen2-VL-7B-Instruct & -0.02 & -0.19 & -0.06 & -0.05 & -0.05 & -0.27 & -0.09 & -0.06 \\
        molmo-72B-0924 & -0.04 & -0.23 & -0.06 & -0.05 & -0.06 & -0.35 & -0.11 & -0.06 \\
        molmo-7B-D-0924 & -0.02 & -0.20 & -0.12 & -0.06 & -0.04 & -0.33 & -0.14 & -0.06 \\
        molmo-7B-O-0924 & -0.04 & -0.26 & -0.11 & -0.05 & -0.05 & -0.34 & -0.10 & -0.06 \\
        molmoE-1B-0924 & -0.02 & -0.22 & -0.13 & -0.04 & -0.04 & -0.35 & -0.17 & -0.05 \\
        \hline
    \end{tabular}
    }
    \caption{PCRI YN scores for different models across additional datasets.}
    \vspace{-1em}
    \label{tab:yn_detail}
\end{table*}

\begin{table*}[ht]
    \centering
    \label{tab:model_comparison_3}
    \renewcommand{\arraystretch}{1.2} % Adjust row height
    \setlength\dashlinedash{0.85pt}    % Dash length
    \setlength\dashlinegap{1pt}       % Gap between dashes
    \setlength\arrayrulewidth{0.8pt}  % Thickness of outer border
    
    % Adjust column width manually
    \scalebox{0.70}{ 
    \begin{tabular}{|l|p{1.8cm}:p{1.8cm}|p{1.8cm}:p{1.8cm}|p{1.8cm}:p{1.8cm}|p{1.8cm}:p{1.8cm}|}
        \hline
        \textbf{Model} & \multicolumn{2}{c|}{\textbf{ChartQA}} & \multicolumn{2}{c|}{\textbf{GQA TestDev Balanced}} & \multicolumn{2}{c|}{\textbf{TextVQA VAL}} & \multicolumn{2}{c|}{\textbf{VizWiz}} \\
        \cline{2-9}
        & PCRI$_{n=2}$ & PCRI$_{n=3}$ & PCRI$_{n=2}$ & PCRI$_{n=3}$ & PCRI$_{n=2}$ & PCRI$_{n=3}$ & PCRI$_{n=2}$ & PCRI$_{n=3}$ \\
        \hline
        InternVL2\_5-1B & 0.202 & -0.207 & 0.063 & -0.099 & 0.190 & -0.266 & 0.112 & -0.122 \\
        InternVL2\_5-26B & 0.264 & -0.159 & 0.080 & -0.115 & 0.266 & -0.215 & 0.121 & -0.149 \\
        InternVL2\_5-2B & 0.260 & -0.210 & 0.078 & -0.143 & 0.263 & -0.276 & 0.128 & -0.175 \\
        InternVL2\_5-4B & 0.286 & -0.175 & 0.086 & -0.077 & 0.257 & -0.221 & 0.123 & -0.083 \\
        InternVL2\_5-8B & 0.302 & -0.142 & 0.076 & -0.116 & 0.282 & -0.179 & 0.127 & -0.103 \\
        Janus-Pro-7B & -0.035 & 0.115 & -0.095 & -1.231 & -0.173 & 0.320 & -0.052 & -2.231 \\
        NVLM & 0.193 & -0.164 & 0.062 & -0.242 & 0.169 & -0.223 & 0.108 & -0.248 \\
        Ovis1.6-Gemma2-9B & 0.251 & -0.262 & 0.090 & 0.131 & 0.189 & -0.322 & 0.155 & 0.211 \\
        Ovis1.6-Llama3.2-3B & 0.166 & -0.251 & 0.070 & 1.000 & 0.105 & -0.333 & 0.135 & 1.000 \\
        Phi-3-Vision & 0.198 & -0.160 & 0.070 & -0.545 & 0.145 & -0.212 & 0.107 & -1.273 \\
        Phi-3.5-Vision & 0.080 & -0.190 & -0.025 & 0.265 & 0.026 & -0.259 & -0.008 & 0.184 \\
        Pixtral-12B & 0.228 & -0.212 & 0.102 & 0.626 & 0.178 & -0.258 & 0.153 & 0.667 \\
        Qwen2-VL-2B-Instruct & 0.243 & -0.189 & 0.075 & -0.030 & 0.190 & -0.265 & 0.119 & 0.006 \\
        Qwen2-VL-72B-Instruct & 0.231 & -0.137 & 0.095 & -0.059 & 0.226 & -0.179 & 0.145 & -0.048 \\
        Qwen2-VL-7B-Instruct & 0.288 & -0.162 & 0.099 & -0.075 & 0.257 & -0.217 & 0.152 & -0.076 \\
        molmo-72B-0924 & 0.217 & -0.199 & 0.111 & -0.020 & 0.167 & -0.255 & 0.156 & 0.010 \\
        molmo-7B-D-0924 & 0.211 & -0.229 & 0.113 & -0.036 & 0.163 & -0.283 & 0.145 & -0.020 \\
        molmo-7B-O-0924 & 0.199 & -0.224 & 0.097 & -0.036 & 0.139 & -0.296 & 0.136 & -0.016 \\
        molmoE-1B-0924 & 0.198 & -0.235 & 0.093 & -0.083 & 0.137 & -0.310 & 0.136 & -0.092 \\
        \hline
    \end{tabular}
    }
    \caption{PCRI VQA scores for different models across additional datasets.}
    \vspace{-1em}
    \label{tab:vqa_detail}
\end{table*}

\begin{table*}[ht]
    \centering
   
    \label{tab:model_comparison_4}
    \renewcommand{\arraystretch}{1.2} % Adjust row height
    \setlength\dashlinedash{0.85pt}    % Dash length
    \setlength\dashlinegap{1pt}       % Gap between dashes
    \setlength\arrayrulewidth{0.8pt}  % Thickness of outer border
    
    % Adjust column width manually
    \scalebox{0.85}{ 
    \begin{tabular}{|l|p{2.5cm}:p{2.5cm}|}
        \hline
        \textbf{Model} & \multicolumn{2}{c|}{\textbf{COCO VAL}} \\
        \cline{2-3}
        & PCRI$_{n=2}$ & PCRI$_{n=3}$ \\
        \hline
        InternVL2\_5-1B & -0.2429 & -0.3047 \\
        InternVL2\_5-26B & -0.4921 & -0.7547 \\
        InternVL2\_5-2B & -0.1009 & -0.1285 \\
        InternVL2\_5-4B & -0.2773 & -0.4005 \\
        InternVL2\_5-8B & -0.3941 & -0.5588 \\
        Janus-Pro-7B & -0.1302 & -0.1613 \\
        NVLM & -0.1665 & -0.3097 \\
        Ovis1.6-Gemma2-9B & -0.1447 & -0.1042 \\
        Ovis1.6-Llama3.2-3B & -0.0932 & -0.0646 \\
        Qwen2-VL-2B-Instruct & -0.0231 & -0.0079 \\
        Qwen2-VL-72B-Instruct & -0.0354 & 0.0082 \\
        Qwen2-VL-7B-Instruct & -0.0492 & -0.0598 \\
        molmo-72B-0924 & -0.1286 & -0.1354 \\
        molmo-7B-D-0924 & -0.1714 & -0.1599 \\
        molmo-7B-O-0924 & -0.0534 & -0.0570 \\
        molmoE-1B-0924 & -0.1304 & -0.1357 \\
        \hline
    \end{tabular}
    }
    \caption{PCRI Caption scores for different models on COCO dataset.}
    \vspace{-1em}
    \label{tab:caption_detail}
\end{table*}

% \begin{equation}
% \frac{P_{\text{patch},\,n}}{P_{\text{whole}}} \approx 1
% \;\Longrightarrow\;
% \mathrm{PCRI}_n \approx 0
% % \qquad (P_{\text{whole}} \neq 0).
% \end{equation}

% $$
% \mathrm{PCRI}_n = 1 - \frac{P_{\text{patch},\,n}}{P_{\text{whole}}}.
% $$

% \begin{equation}
% \mathrm{cPCRI}_n \;=\; 1 \;-\; \frac{P_{\text{patch},\,n}-C}{P_{\text{whole}}-C},
% \qquad P_{\text{whole}}\neq C .
% \end{equation}

% $$
% \frac{P_{\text{patch},\,n}}{P_{\text{whole}}} \approx 1 \;\Longrightarrow\; \mathrm{PCRI}_n \approx 0.
% $$

% $$
% \frac{P_{\text{patch},\,n}}{P_{\text{whole}}} \approx 1 \;\Longrightarrow\; \mathrm{PCRI}_n \approx 0, \quad P_{\text{whole}}\neq 0.
% $$

\end{document}